%% file: main.tex
\documentclass[sigconf,nonacm]{acmart}

\usepackage[]{hyperref}

\usepackage{mathtools}
\usepackage{amsmath}

\usepackage[group-separator={,},per-mode=symbol]{siunitx}

\usepackage[disable]{todonotes}

\usepackage{algorithm}
\usepackage{algpseudocode}

\usepackage{graphicx} %

\usepackage{tikz}
\usepackage{subcaption}
\usetikzlibrary{patterns}
\usepackage{xcolor}

\usepackage{threeparttable}
\usepackage{multirow}
\usepackage{makecell}
\usepackage{pifont}

\usepackage[column=O]{cellspace}

\usepackage{pgfplots}
\usepackage{graphicx}
\pgfplotsset{compat=1.18}

\usepackage{comment}

\AtBeginDocument{%
  }

\begin{document}

\title{SlimPipe: Memory-Thrifty and Efficient Pipeline Parallelism for Long-Context LLM Training}

\author{
{\rm Zhouyang Li$^*$, Yuliang Liu$^{*\dag}$, Wei Zhang, Tailing Yuan,} \\
 \rm{Bin Chen, Chengru Song, and Di Zhang}\\[6pt]}
\thanks{$^*$Equal contribution.}
\thanks{$^\dag$Corresponding author: Yuliang Liu <liuyuliang@kuaishou.com>}

\affiliation{%
        \institution{Kuaishou Technology}
        \city{Beijing}
        \country{China}}

\renewcommand{\shortauthors}{Zhouyang Li et al.}

\begin{abstract}
Pipeline Parallelism (PP) serves as a crucial technique for training Large Language Models (LLMs), owing to its capability to alleviate memory pressure from model states with relatively low communication overhead. However, in long-context scenarios, existing pipeline parallelism methods fail to address the substantial activation memory pressure, primarily due to the peak memory consumption resulting from the accumulation of activations across multiple microbatches. Moreover, these approaches inevitably introduce considerable pipeline bubbles, further hindering efficiency.

To tackle these challenges, we propose \textit{SlimPipe}, a novel approach to fine-grained pipeline parallelism that employs uniform sequence slicing coupled with one-forward-one-backward (1F1B) schedule. It reduces the accumulated activations from several microbatches to just one, which is split into several slices. Although the slices are evenly partitioned, the computation cost is not equal across slices due to causal attention. We develop a sophisticated workload redistribution technique to address this load imbalance. SlimPipe achieves (1) \textit{near-zero memory overhead} and (2) \textit{minimal pipeline bubbles} simultaneously. The effectiveness of SlimPipe has been proven by thorough testing with diverse model architectures, context window sizes, and SlimPipe-specific configurations.
For example, compared to state-of-the-art methods, SlimPipe significantly boosts the Model FLOPs Utilization~(MFU) to up to $1.57\times$. More notably, on Llama~70B with a context length of 2048K, it maintains over 45\% MFU on 256 NVIDIA Hopper 80GB GPUs.

\end{abstract}

\maketitle

\section{Introduction}

\input{img/mem_comparison.tex}
\input{img/mem_pp_rank.tex}

Autoregressive large language models, such as Llama~\cite{touvron2023llama1, touvron2023llama2}, Gemini~\cite{geminiteam2024geminifamilyhighlycapable}, GPT~\cite{brown2020language},  and Mixtral~\cite{jiang2024mixtral}, have achieved dominant performance in many NLP tasks. At the same time, an increasing number of studies have found that context length plays a significant role in model applications, such as multi-turn dialogue and long-context understanding.
However, as the model size and context length continue to grow, the storage requirements far exceed the capacity of modern accelerators. This memory demand comprises two primary components: (1)~\textit{model states}, including parameters, gradients, and optimizer states, which scale proportionally with model size, and (2)~\textit{activations}, whose memory footprint grows linearly with context length. 

Pipeline Parallelism has become an indispensable component in hybrid parallelization strategies for training LLMs due to its ability to mitigate memory pressure from model states while maintaining relatively low communication overhead. However, existing PP approaches exhibit significant limitations in handling the memory pressure caused by activations, which scales with context lengths. As illustrated in Figure~\ref{fig:memory_comparison}, while the memory footprint of the model states decreases as the size of the PP increases, the activation memory remains constant. This limitation poses substantial challenges for exploring long-context models. For instance, as shown in Figure~\ref{fig:max_len_comparison}, even with maximal TP utilization within NVLink domains, the maximum viable context length for Llama~13B models is limited to 124K. Further context length expansion requires either memory-computation trade-offs through activation recomputing or sequence partitioning across nodes with low communication bandwidth, both of which significantly compromise training efficiency.

Beyond memory constraints, pipeline bubbles present another critical challenge in long-context scenarios. Both the 1F1B schedule~\cite{narayanan2021memory, fan2021dapple} and its interleaved variant introduce \textit{warm-up bubbles} due to the necessity of forward warm-up phases to fill the pipeline and corresponding backward cool-down phases before synchronization. In long-context settings, limited global batch sizes exacerbate the relative impact of these warm-up bubbles. ZB-V~\cite{qi2024pipelineparallelismcontrollablememory, qi2024zero} addresses this issue by decoupling the computations of the activation gradient ($T_b$) and the weight gradient ($T_w$) in the backward pass. When $T_f=T_b=T_w$, where $T_f$ represents the forward computation time, ZB-V can achieve zero bubble. However, the time differences among $T_f$, $T_b$, and $T_w$ can be substantial. For instance, the core attention operation, which lacks weights, implies $T_w=0$, while $T_b$ is also significantly larger than $T_f$. This difference introduces \textit{imbalance bubbles}. The issues related to the imbalance bubbles of ZB-V will be thoroughly analyzed in the background and related work section. The current pipeline bubble characteristics are illustrated in Figure~\ref{fig:bubble_comparison}.

The fundamental limitation stems from the inherent nature of the classic PP design. Our key observation is that to achieve optimal throughput, the pipeline must accumulate $p$ computational units during the warm-up phase, where $p$ is the PP size. However, since existing PP methods operate at \textbf{\textit{microbatch-level}} granularity, these methods require storing $p$ microbatches of activations simultaneously. Although the number of layers per pipeline stage is reduced by a factor of $p$, the activation memory consumption remains constant due to the accumulation of $p$ microbatches. Inspired by TeraPipe~\cite{li2021terapipe}, we have developed a fine-grained 1F1B schedule that decomposes the conventional microbatch-level computing units into finer \textbf{\textit{slice-level}}, where each slice represents a sliced segment of a microbatch. To maintain a stable memory footprint during the steady phase, we evenly divide the sequence into equal-length slices. This innovation enables us to minimize memory overhead, including those arising from accumulated activations, as depicted in Figure~\ref{fig:memory_comparison}.

However, because mainstream LLMs employ causal attention, even though the slices are of equal length, the later slices impose higher computational cost than the earlier ones. This leads to pipeline imbalance bubbles. To counteract this, we design a novel method that equalizes memory usage and computational workload across different slices by redistributing the attention workload among different pipeline parallelism devices.
The load imbalance with uniform slicing is primarily due to the attention calculation, so we redistribute an appropriate portion of the attention operation from heavily loaded devices to those with lighter load. This redistribution enables the less burdened devices to handle the additional attention computations and then send the outcomes back to the original devices.
By employing this method, we maintain an equilibrium of computational and memory load across all uniform slices. This means that imbalance bubbles are completely eliminated.
Moreover, in LLM training, to maintain model performance, the global token size is constrained by the critical batch size~\cite{mccandlish2018empirical}. Longer sequences reduce the batch size and inflate the pipeline bubble overhead. By operating at slice-level granularity instead of microbatch-level, we can significantly reduce the pipeline warm-up bubbles. To summarize, we address pipeline bubble issues through two key mechanisms: (1)~context redistribution that eliminates imbalance bubbles, and (2)~fine-grained scheduling that reduces warm-up bubbles, collectively achieving near-zero pipeline bubbles, as depicted in Figure~\ref{fig:bubble_comparison}.

To this end, we introduce \textit{SlimPipe}, a fine-grained pipeline parallelism scheme that employs uniform sequence slicing combined with 1F1B scheduling. SlimPipe delivers three key advantages: (1)~near-zero memory overhead, (2)~minimal pipeline bubbles, and (3)~enhanced computational efficiency by leveraging its memory-thrifty design to minimize both activation recomputing and parallelism size requirements. Our evaluations on various models and context lengths demonstrate that SlimPipe can boost the LLM training throughput to up to $1.57\times$ compared to state-of-the-art systems. In summary, our paper presents the following contributions:
\begin{enumerate}%
\item Our analysis reveals two fundamental limitations in current pipeline parallelism approaches: memory inefficiency and significant pipeline bubble overhead. Moreover, we demonstrate that these issues exacerbate their difficulties in long-context training.

\item We propose SlimPipe, a novel approach to fine-grained pipeline parallelism that employs uniform sequence slicing coupled with 1F1B scheduling.

\item We propose a novel workload redistribution technique. By dynamically reallocating the attention workload between PP devices handling different slices, we achieve a balanced distribution of both computational and memory load among them.

\item Our extensive evaluation demonstrates that SlimPipe consistently outperforms state-of-the-art systems, such as DeepSpeed~\cite{jacobs2023deepspeed} and Megatron-LM~\cite{shoeybi2019megatron, narayanan2021efficient}, achieving significant improvements in both memory utilization and throughput.
\end{enumerate}

\section{Background and Related Work}

The notations used in this paper are listed in Table~\ref{tab:notations}.

\begin{table}[t]
  \centering
  \small
  \caption{Notations with description.}
  \begin{tabular}{cl}
    \toprule
    Notation & Description \\
    \midrule
    $L$   & number of transformer layers \\
    $M_h$ & size of one embedding tensor \\
    $M_a$ & total size of activations per microbatch \\
    \midrule
    $t$ & tensor parallelism size       \\
    $c$ & context parallelism size      \\
    $d$ & data parallelism size         \\
    $e$ & expert parallelism size       \\
    $p$ & pipeline parallelism size     \\
    \midrule
    $m$ & number of microbatches               \\
    $n$ & number of slices per input sequence  \\
    $v$ & number of stages per pipeline device \\
  \bottomrule
  \end{tabular}
  \label{tab:notations}
\end{table}

\subsection{Intra-operator Parallelism}

\noindent\textbf{Data Parallelism} (DP)~\cite{7239545} involves replicating the model across multiple devices, with each device processing a different subset of the data. The gradients are then aggregated across all devices to update the model parameters. Data parallelism is effective for scaling out the training process by distributing the dataset.

{\setlength{\parskip}{3pt}

\noindent\textbf{Tensor Parallelism} (TP)~\cite{shoeybi2019megatron} is achieved by partitioning the parameters and inputs within an operator, effectively splitting the computational burden across multiple devices. This is particularly useful when the model is too large to fit into the memory of a single device. However, due to the significant communication overhead of tensor parallelism, it is generally confined within a node. \textbf{Sequence Parallelism} (SP)~\cite{korthikanti2023reducing}, as explained in the Megatron-LM framework, ensures that data remains in a partitioned state throughout the model training process, rather than maintaining redundant activations outside the attention and MLP modules.

\noindent\textbf{Context Parallelism} (CP) is an advanced technique used to train LLMs by distributing the sequence across multiple devices. \textit{DeepSpeed Ulysses}~\cite{jacobs2023deepspeed} is a system optimization designed to enable the training of extremely long-context transformer models. \textit{Ring Attention}~\cite{liu2023ring,brandon2023striped} is an approach that leverages blockwise computation of attention to distribute long sequences across multiple devices. 

\noindent\textbf{Expert Parallelism} (EP)~\cite{gale2023megablocks, hwang2023tutel, lepikhin2020gshard, rajbhandari2022deepspeed} is a technique utilized in Mixture-of-Experts (MoE)~\cite{fedus2022switch, zoph2022st, jiang2024mixtral} architectures. In an MoE layer, different experts are dynamically activated according to the input tokens. EP distributes these specialized experts across multiple devices, enabling efficient computation by activating only the relevant experts for each input.
}  %

\subsection{Pipeline Parallelism}
Pipeline Parallelism~\cite{kim2023bpipe, huang2019gpipe,narayanan2019pipedream, narayanan2021memory, li2021terapipe, fan2021dapple, narayanan2021efficient, qi2024zero, qi2024pipelineparallelismcontrollablememory} splits a model across its layers, assigning distinct layers to different devices for simultaneous processing. However, this method can result in bubbles, which are caused by uneven workload distribution or periods of inactivity while waiting for gradient synchronization. To mitigate the problem of bubbles in pipeline parallelism, several studies have investigated solutions focusing on asynchronous updates. However, it carries the potential risk of affecting model performance. In this paper, we only consider synchronous pipeline parallelism .

\input{table/compare_sched}

\noindent\textbf{Memory Consumption.} GPipe~\cite{huang2019gpipe} requires accumulating the activations for all microbatches until the backward pass is completed for the first microbatch, leading to high peak memory consumption. Terapipe~\cite{li2021terapipe} exploits the autoregressive nature of LLMs through its token-level scheduling, but it inherits GPipe's critical memory limitation: accumulating all activations throughout the pipeline.

PipeDream-Flush~\cite{narayanan2021memory} and DAPPLE~\cite{fan2021dapple} introduces the 1F1B pipeline schedule, which divides each iteration into three phases: warm-up phase, steady phase, and cool-down phase. The warm-up phase aims to bring all pipeline devices into a working state. To ensure the availability of backward operations, activations generated during the warm-up phase are accumulated. Subsequently, all pipeline devices enter the 1F1B state during the steady phase, where the number of activations consumed by backward is equal to the number of activations produced by forward, thus maintaining stable device memory usage. This results in an accumulation of $p$ microbatch activations. Although the 1F1B schedule eliminates the dependence of activation accumulation on $m$, it fails to achieve inverse scaling with increasing $p$. In contrast, SlimPipe attains \textbf{near-zero memory overhead}, with activation memory consumption scaling inversely with $p$.

Zero Bubble Pipeline Parallelism~\cite{qi2024pipelineparallelismcontrollablememory, qi2024zero} introduces a novel V-shape pipeline schedule that effectively balances memory utilization across pipeline devices. Through the V-shape schedule, ZB-V maintains the same peak memory overhead as 1F1B, while V-Half and V-Min reduce the peak memory to 1/2 and 1/3 of that of 1F1B, respectively. However, the activation memory consumption remains a constant that does not decrease with increasing $p$.

{\setlength{\parskip}{3pt}
\noindent\textbf{Bubble Fraction.}}
Gpipe treats the entire computation of a microbatch across all layers in a pipeline stage as an atomic \textit{computation unit}. This coarse granularity results in the pipeline stages experiencing significant warm-up bubbles. 

To address the issue of bubble overhead, several works have approached the problem from both the model and input data perspectives. Instead of treating the entire batch as the computation unit, TeraPipe breaks down the computation into smaller units at the token level. This approach allows for more fine-grained scheduling and reduces the idle periods between pipeline stages. On the other hand, interleaved 1F1B~\cite{narayanan2021efficient} builds upon the 1F1B schedule and approaches the problem from the model perspective. In interleaved 1F1B, each device can undertake computations for several discrete subsets of layers rather than a single continuous sequence of layers. SlimPipe synergistically combines fine-grained model and input data sharding to \textbf{minimize warm-up bubble overhead}. To address workload imbalance caused by causal attention among slices, we propose a novel context redistribution mechanism that dynamically balances computational load across pipeline stages, effectively \textbf{eliminating imbalance bubbles}.

ZB-V~\cite{qi2024zero} innovatively splits backward into two parts: one for input gradient and the other for parameter gradient, theoretically eliminating bubbles. However, assuming the forward, input gradient, and parameter gradient computation costs for a microbatch are $T_f$, $T_b$, and $T_w$, respectively, bubbles can only be avoided if $T_f = T_b = T_w$, which is impossible for Transformer models with attention blocks. For attention block, theoretically, $T_b = 2 \times T_f \gg T_w=0$. When accounting for modern optimizations like Flash Attention and the inherent MFU disparity between forward/backward passes, the situation further deteriorates. Given that the computational complexity of attention is quadratic with respect to context length, the attention component tends to dominate the entire training process in long-context scenarios. This leads to significant challenges for ZB-V in long context training. 

In summary, SlimPipe outperforms other pipeline parallelism methods in terms of memory efficiency, achieving near-zero memory overhead. It also demonstrates excellent performance in reducing pipeline bubbles. The results are presented in Table~\ref{tab:sched-comparison}. It is worth emphasizing that in long-context training, SlimPipe's advantages become even more pronounced.

\subsection{Activation Rematerialization}
\textbf{Activation Checkpointing}~\cite{chen2016training, kirisame2021dynamic} is a memory optimization technique that reduces memory usage by strategically releasing activation. When applied across a sequence of layers, this method retains only the initial layer's input for the backpropagation. During the backward pass, intermediate outputs are recalculated as needed for gradient computation. This strategy significantly reduces the memory footprint of activations, thereby freeing up memory resources to accommodate larger models. Various instruments~\cite{Beaumont2019OptimalCF, jain2020checkmate, patil2022poet} automate the decision-making process, fine-tuning strategies to optimize runtime efficiency while staying within a predefined memory limit. Meanwhile, a recent work~\cite{yuan2024accelerating} introduces the concept of the Pareto frontier for activation checkpointing strategies in the context of LLM training. According to this approach, the strategy is developed by training models along the Pareto frontier, optimizing the trade-off between memory consumption and runtime efficiency.

{\setlength{\parskip}{3pt}

\noindent\textbf{Activation Offloading}~\cite{shriram2019dynamic, rhu2016vdnn}, similar to activation checkpointing, is another memory optimization technique. It achieves memory savings by swapping activations that are temporarily not needed from device memory to host or NVMe memory, and then swapping them back to device memory when they are required.

\input{img/slim_1f1b.tex}

\section{Challenges in Long Context LLM Training}\label{sec:challenges}
}  %
{\setlength{\parskip}{3pt}
\noindent\textbf{Immense Memory Overhead.} Current training of long-context LLMs faces substantial memory pressure, which drives the need for hybrid parallelism. The memory pressure consists of model states due to the large model size and activations due to the long context. Taking a 70B parameter Llama model and 1M context length as an example, even when using full recomputing with $t = 8$, the total activation memory consumption is \SI{160}{\gibi\byte}, which far exceeds the \SI{80}{\gibi\byte} memory limit of mainstream GPUs. The calculation is as follows: 1048576\,(context length)\,$\times$ 8192\,(hidden dimension)\,$\times$ 80\,(number of layers)\,$\times$ 2\,(bytes per \textsl{bfloat16})\,$\div$ 8\,(tensor parallelism size).

\noindent\textbf{Limited Global Batch Size.} Long context LLMs training maintains a relatively fixed global tokens size per iteration. This practice is to ensure the model performance. The concept of critical batch size has been introduced by~\cite{mccandlish2018empirical}. Exceeding this size by inflating the global tokens size can result in excessive noise reduction, thereby adversely affecting the model's performance. This means that as
the context length continues to increase, the global batch size will
scale inversely. The reduction in global batch size directly decreases the number of microbatches ($m$), as evidenced by Table~\ref{tab:sched-comparison}, which proportionally increases the warm-up bubble overhead.

\noindent\textbf{Imbalanced Model Partition.} The vocabulary size is essentially large in recent LLMs.
For example, the Llama~3~\cite{dubey2024llama3herdmodels} models have a vocabulary size of \num{128000},
and the Gemma~\cite{team2024gemma} models even have one of \num{256128}.
Most PP schemes simply assign the vocabulary to the last pipeline device.
This sort of uneven model partitioning not only brings considerable memory overhead, but also creates substantial workload imbalance among PP devices.
}  %

\section{Methods}
In this section, we delve into the particular methods used by SlimPipe.
First, we introduce how fine-grained pipeline parallelism with \textit{uniform slicing} saves both memory and warm-up bubble time.
We also provide a theoretical analysis to reveal the characteristics of our newly designed pipeline scheme.
Next, we discuss the issue of workload imbalance among pipeline devices due to uniform slicing and causal attention.
After that, we demonstrate how \textit{attention context exchange} addresses the issue of workload imbalance.
Finally, we employ \textit{vocabulary parallelism} to equally distribute the computation and memory of the output layer among pipeline devices.

\subsection{Pipeline Schedule with Uniform Slicing}
\input{img/interleaved_slim_1f1b}
\input{img/memory_red.tex}
\subsubsection{Unifrom Slicing}\label{sec:slicing}
Pipeline Parallelism~\cite{huang2019gpipe, narayanan2019pipedream} accumulates several forward passes on each device to warm up the pipeline.
These prefilled passes bring significant activation memory, especially on lower-rank devices of the pipeline.
The long input sequence could be split into short slices to reduce the activation memory of each forward pass.
And the straightforward approach to split a sequence is uniform slicing.
However, the equal slice length will lead to unequal computation time between slices.
The reason is the causal attention mechanism~\cite{li2021terapipe}, by which a latter query attends to former keys and values, but not vice versa.
We will discuss and solve this issue later in Section~\ref{sec:context_red}.
Despite that issue, uniform slicing has the following substantial advantages over non-uniform slicing:
\begin{enumerate}%
\item The maximum amount of accumulated memory is better constrained.

\item The fixed slice length makes it more feasible for other techniques (e.g., context parallelism) to divide at the sequence dimension.

\item Slices are prevented from being too short to maintain sufficient arithmetic intensity.
\end{enumerate}%

\subsubsection{Slice-wise Scheduling}
Firstly, we split the input sequence into $n$ equal-length slices, where $n$ is a multiple of $p$.
To effectively calculate the causal attention slice by slice, we deploy the KV cache technique~\cite{pope2023efficiently}.
Remarkably, the KV cache imposes no memory overhead on the accumulated activation.
Because the keys and values are deliberately retained for gradient calculation in backward passes.

Next, we apply uniform slicing to the 1F1B pipeline scheme~\cite{narayanan2019pipedream}.
The 1F1B schedule is adopted with some vital adjustments.
To perform backpropagation~\cite{rumelhart1986learning} for slices in one sequence, respectively, their backward passes are scheduled in reverse order of forward passes.
Paired with this last-in first-out schedule, we release the corresponding portion of KV cache immediately once a backward pass is finished.
This ensures that the crucial invariant of maximum accumulated memory is maintained through the 1F1B steady phase.
In addition, we put more forward passes ahead to align forward and backward passes separately.
The crafted pipeline scheme is depicted in Figure~\ref{fig:slim_schedule}.

Furthermore, we smoothly extend our pipeline scheme to an interleaving form~\cite{narayanan2021efficient},
by assigning multiple small stages to each device.
As illustrated in Figure~\ref{fig:interleaved_slim_schedule}, uniform slicing and stage interleaving synergize well with each other.
The accumulated activations and warm-up bubbles are further reduced.
\subsubsection{Theoretical Analysis}\label{sec:analysis}
As annotated in Figure~\ref{fig:slim_schedule},
although the number of forward passes in the warm-up phase is increased,
the total accumulated activation is actually decreased.
Given that $n$ is equal to or greater than $p$,
the total accumulated activation size
\begin{equation}
  M_{acc} = (1+\delta)\frac{M_a}{p}, \quad\delta=\frac{2(p-1)}{n}
\end{equation}
is much smaller than $M_a$.
It is even approaching $M_a/p$ as $n$ continues increasing,
as if virtually divided by the PP size.
Moreover, the bubble time during warm-up and cool-down phases is reduced by about $n$ times.
In fact, the bubbles shrink super-linearly due to the causal attention mechanism.
As a result, the bubble fraction is less than $\displaystyle \frac{p-1}{nm}$.
When attention dominates the computation time (with an extremely long context length), the bubble fraction will become $\displaystyle \frac{(p-1)p}{(n+1)nm}$.
Besides these benefits, the overall communication volume of default 1F1B has not been changed.
The two significant characteristics of SlimPipe are depicted in Figure~\ref{fig:memory_dec}.
With a fairly large value of $n$, the activation memory and bubble time are substantially reduced.
In terms of the interleaving form, these two attributes will be further reduced by the number of stages $v$ (formulas are listed in Table~\ref{tab:sched-comparison}).

\subsection{Attention Context Exchange} \label{sec:context_red}
\input{img/imba.tex}
\input{img/attn_balance_2.tex}
\subsubsection{Imbalance Bubbles}
If we just split each sequence into equal-length slices, the execution timeline will not be as ideal as depicted in Figure~\ref{fig:interleaved_slim_schedule}.
Although we try to align each kind of passes across devices, bubbles arise and pervade between them, which is roughly sketched in Figure~\ref{fig:imba_bubble}.
These bubbles are caused by the uneven computation time of causal attention.
As the query length is fixed by uniform slicing, the computation time is in proportion to the length of attended key-value.
Latter slices, attending to more KV cache, cost more time to compute in both forward and backward passes.

At a specific moment, the workloads across pipeline devices conform to an arithmetic progression.
And the difference between the heaviest and the lightest is $p-1$ slices of key-value.
The situation becomes worse at the juncture of two microbatches, where the workload difference could be as great as $n-1$.
Devices processing with less KV cache finish their computation sooner, and have to wait for other devices to communicate with them.
This waiting-time is a type of imbalance bubbles.
Imbalance bubbles will propagate along pipeline devices in both directions and cross with each other, making the pipeline timeline quite irregular.

\subsubsection{Exchange the Context}
To eliminate these messy imbalance bubbles, we need to make the attention workload as equal as possible across devices.
The essential condition is that all devices at one moment have equal computation time.
Note that it does not require a uniform computation time across passes, which is otherwise required by TeraPipe.
This essential condition allows us to redistribute the attention workload on the spot.
As depicted in Figure~\ref{fig:attn_balance},
a device with more KV cache sends its query and portions of key-value to a device with less KV cache,
and then the attention will be calculated there.
The remotely calculated attention output will be sent back and merged with the locally calculated one via the online softmax method~\cite{milakov2018online}.
The attention workloads after redistribution become almost equal. The difference between them is at most one slice of key-value.

\subsubsection{Communication Volume}
We use the non-interleaving form of SlimPipe to demonstrate the communication volume of context exchange.
The exchanged context is $1$ slice of query (Q) and output (O) plus $\lfloor(p-1)/2\rfloor$ slices of key (K) and value (V),
and the latter becomes $\lfloor(n-1)/2\rfloor$ slices at the microbatch juncture.
The juncture period occupies $p-1$ out of $n$ passes per microbatch.
Generally, the unsliced query, key, value, and output all have the size $M_h$.
So one slice of each from one device has the size $\displaystyle \frac{L}{p}\cdot\frac{M_h}{n}$.
Finally, the total volume of exchanged context per microbatch per device is
\begin{equation} \label{eq:cr}
  \begin{split}
    \Theta &= \Bigl(\underbrace{2n}_{\text{Q+O}}+\underbrace{2(n-p+1)\Bigl\lfloor\frac{p-1}{2}\Bigr\rfloor}_{\text{K+V, not at junctures}}+\underbrace{2(p-1)\Bigl\lfloor\frac{n-1}{2}\Bigr\rfloor}_{\text{K+V, at junctures}}\Bigr)\frac{LM_h}{pn} \\
                      &\le \Bigl(2 - \frac{p-1}{n}\Bigr)LM_h
  \end{split}
\end{equation}
The inequality originates from the round-down operation.
This volume is at most $2LM_h$, virtually independent from the PP size and number of slices.
When applied to the interleaving form, context exchange gets a communication volume at the same degree.

\subsection{Vocabulary Parallelism}
\subsubsection{The Output Layer}
Although we have balanced the memory consumption and computation workload in the previous sections,
the pipeline parallelism still suffers from another sort of imbalance.
The output layer of a language model is a GEMM operation that projects hidden states into the vocabulary space.
As described in Section~\ref{sec:challenges}, the vocabulary size is really large.
The output layer is calculated by the last pipeline device, thus significantly aggravating the workload imbalance here.
As depicted in Figure~\ref{fig:post_bubble}, this workload imbalance also causes bubbles in the middle of the pipeline.
Moreover, when training LLMs, the output layer is followed by a cross-entropy loss, which will store the vocabulary logits in \textsl{float32} for gradient calculation.
That brings a large activation footprint to the last PP device.
For instance, with a context length of 256K and a vocabulary size of \num{128000}, it
consumes about \SI{16}{\gibi\byte} of GPU memory even in 8-way TP. 

\subsubsection{Distribute the Vocabulary}
The point is to equally distribute the computation and memory of the output layer among pipeline devices.
Specifically, we parallelize the GEMM along its vocabulary dimension (column-wise).
The input hidden states are broadcasted to all PP devices.
Thus, each device can calculate a part of the GEMM and get a portion of the vocabulary logits.
Rather than gathering these portions together,
we calculate the cross-entropy loss from the sharded logits, and the necessary statistics are synchronized.
The scalar statistics are much smaller than the logits, so the communication volume is drastically reduced.
Since the word embedding and the output layer usually share weights~\cite{press2016using, vaswani2017attention}, we also parallelize the former.
This method was first adopted by the TP~\cite{shoeybi2019megatron} implemented in Megatron-LM.
However, it could not be trivially applied to common PP schemes due to their misaligned schedules.
Since all passes are tidily aligned by attention context exchange, SlimPipe is particularly suitable for this method.
\input{img/post_bubble.tex}

\section{Implementation}\label{sec:impl}
We build SlimPipe within Megatron-LM~\cite{shoeybi2019megatron, narayanan2021efficient}, %
and implement our functions using PyTorch~\cite{paszke2019pytorch}. %
We make some practical efforts to reduce the per-layer activations.
We use the cuDNN~\cite{chetlur2014cudnn} Scaled Dot Product Attention (SDPA) as the primitive of our attention layer.
cuDNN SDPA is similar to FlashAttention~\cite{dao2022flashattention}, which avoids storing large intermediate matrices for backward.
Our SwiGLU implementation recomputes the swish function instead of storing the intermediate activations.
We also adopt a memory-efficient RMSNorm, which otherwise uses its output to calculate gradients.
Moreover, we customize the techniques as below, to better utilize them in SlimPipe.

{\setlength{\parskip}{3pt}
\noindent\textbf{Chunked KV Cache.} The size of the KV cache grows up along the forward passes and shrinks down along the backward passes.
We store the KV cache in a list of individual tensors (chunks), instead of joining them together into an entire tensor.
The joined tensor would occupy a large contiguous buffer, which will be frequently re-allocated, leading to severe memory fragmentation in the allocator.
While PagedAttention~\cite{kwon2023efficient} stores the KV cache in relatively small blocks, we store them in slice-sized chunks
to avoid block addressing in the GPU kernel of attention.
This is enough to eliminate memory fragmentation since all chunks have identical size by uniform slicing.
These chunks will be precisely reused between two adjacent microbatches in the pipeline,
where the backward pass releases one and the forward pass acquires one.

\noindent\textbf{Early Key-Value Exchange.} Attention context exchange involves additional communication among pipeline devices.
The communication costs may offset the benefit from the balanced workload. 
However, we do not need to send the key-value of the last few slices,
as illustrated in ~\ref{fig:attn_balance}.
Instead we could just send those of the first few slices that are produced earlier.
Thereby we can communicate them much earlier than the attention calculation and even overlap it with the computation of previous slices.
By this optimization, we substantially alleviate the delay in attention calculation.

\noindent\textbf{Commutated Context Parallelism.}
Context parallelism is widely utilized to train LLMs with extremely long context.
Existing CP implementations %
communicate key-value among devices and use them to calculate the attention output against the local query.
When CP is used along with KV cache, the cached key-value will be communicated every time a later slice comes, which is rather inefficient.
To eliminate the redundant communication, we implement a commutated variant of context parallelism.
This CP variant communicates the query, output and the normalizer in softmax instead of the key and value.
In general, the query and output have the same size as the key and value, and the scalar normalizer is too small to count.
Therefore, the communication volume of CP is recovered to that without KV cache. %
}
\section{Evaluation}

\subsection{Experimental Settings}
\input{table/model_config.tex}
All experiments are conducted on a GPU cluster where each node is configured with two Intel Xeon Platinum CPUs and 1TB of memory.
In each node, 8 NVIDIA Hopper 80GB GPUs are installed and interconnected via NVLink, whose bandwidth is \SI{400}{\giga\byte\per\second} per GPU.
Besides that, every GPU is coupled with a \SI{400}{Gbps} NIC for inter-node communication.
Unless otherwise stated,
TP, CP and EP should be deployed within a node, while PP and DP could be deployed across nodes.
TP is always paired with SP.

Models of various architectures are used in evaluation.
We select the Llama-like dense models and the Mixtral series MoE models.
The specifications of the models are listed in Table~\ref{tab:model_config}.
GQA~\cite{ainslie2023gqa} is applied to models except Llama~13B.
In MoE layers, 2 out of 8 experts are routed for each token.
The expert router is set to complete balance for performance measurement.
Every model is coupled with a vocabulary of \num{128000} entries.
The training precision is \textsl{bfloat16} except that \textsl{float32} is used in loss calculation and gradient accumulation.
The optimizer is Adam with \textsl{float32} internal states.

\subsection{Memory Saving}
\input{data/memory_by_p.tex}
We use a simple experiment to show how SlimPipe reduces the memory usage in training.
We train the Llama~13B model with input sequence lengths of 32K, 64K and 96K.
The TP size is 8 and the PP size increases from 2 to 8.
The number of stages per device is set to $L/p$ (i.e., maximum interleaving stages).
We measure the memory usage of the first and the last pipeline devices, using the \texttt{torch.cuda.max\_memory\_allocated} method.

We draw the measured data as markers in Figure~\ref{fig:mem_by_p}.
We also draw a curve for each sequence length, whose formula is $M_t/p$,
where $M_t$ is the theoretical memory required to train the model without PP (only with 8-way TP).
We use the memory modeling method from previous works~\cite{korthikanti2023reducing,yuan2024accelerating}.
As we can see, the memory usage for both devices decreases in close alignment with the theoretical curves.
The total memory is nearly in an inverse proportional relationship to the PP size, as if distributed by PP.
While classic PP distributes model states, SlimPipe additionally distributes activations.
In particular, the activations of the output layer are also distributed, as evidenced by our measurements on the last device.
The memory usage of the first device is slightly higher than that of the last device.
The gap between them is formulated as $2(p-1)M_a/nvp$.
We can claim that nearly all the memory used in LLM training is distributed in PP.
The only existing technique that can achieve this is TP (with SP), but at significantly higher communication costs.

\subsection{Insights into Slice Length}
\input{data/tps_vs_n.tex}
As discussed in Section~\ref{sec:slicing},
fine-grained slicing can reduce the accumulated activations but also decrease the arithmetic intensity.
So in this experiment, we will study how the value of $n$ affects the training efficiency,
and find an appropriate $n$ for a given context length.
We train the Llama~13B model with 2 microbatches per iteration.
The three sets of input sequence lengths are 128K, 256K and 512K respectively.
The parallelism configuration is 8-way TP combined with 4-way PP, and full checkpointing is enabled.
The number of stages per pipeline device is 5,
and the number of slices per sequence gradually increases from $p$ to $8p$.
We use MFU to quantify the training efficiency.

As we can see in Figure~\ref{fig:tgs-comparison},
fine-grained slicing improves MFU initially (due to reduced bubbles) but harms it eventually (due to lower arithmetic intensity) as slices become shorter.
The MFU of 128K context length drops sharply after $n=2p$, indicating that the arithmetic intensity is too weak.
We also see that the transition point of $n$ comes later for a longer context length.
MFU of the 512K context length remains high even with 32 slices per sequence.
In conclusion, we should choose a moderate value of $n$ to maintain the training efficiency while saving memory.

\input{img/system.tex}
\subsection{End-to-End Performance}\label{sec:system}
When SlimPipe is integrated into a hybrid parallelism training system,
it could introduce new optimization opportunities to other techniques.
For example, the GPU memory saving could avoid the use of full checkpointing, thus increasing the system efficiency.
In the main course, we compare the system performance between Megatron-LM and SlimPipe.
For clarity, we refer to the baseline system with interleaved 1F1B PP as Megatron-LM, and the system with our innovative PP as SlimPipe.
Prevalent parallelism techniques (TP, CP, EP, DP, and PP) are organically composed in both systems.
The activation saving techniques mentioned in Section~\ref{sec:impl} are applied uniformly,
while full or selective checkpointing~\cite{korthikanti2023reducing,yuan2024accelerating} is enabled on their demand.
The selective checkpointing implemented by us just recomputes the up projection plus SwiGLU in an MLP layer.
DeepSpeed~\cite{rasley2020deepspeed} is also involved and is equipped with its Zero Redundancy Optimizer~\cite{rajbhandari2020zero} and Ulysses Parallelism (UP).
To exhibit the best performance of each system, their hybrid parallelism configurations are baked through grid search.

We select 4 models covering dense and MoE architectures, in various sizes.
The input sequence length extends from 64K to 512K.
The per-iteration token size is fixed to 4M, which means that there are fewer microbatches when the sequence length is longer.
The total number of GPUs scales from 128 to 512.
The comprehensive benchmark results are shown in Figure~\ref{fig:system}.
SlimPipe outperforms the other two systems in training efficiency through all models, context lengths, and number of GPUs.

{\setlength{\parskip}{3pt}
\noindent\textbf{Difference in Context Lengths.} SlimPipe demonstrates increasingly significant advantages when training with longer context lengths. Extended contexts inherently demand greater activation memory consumption and incur more significant warm-up bubbles. As previously discussed, SlimPipe's fine-grained schedule minimizes bubbles overhead, thereby delivering substantial efficiency benefits. Moreover, leveraging its memory-thrifty design, which substantially reduces activation storage requirements, we maintain training efficiency without resorting to full checkpointing or expensive cross-node TP/CP communication. In contrast, the other two approaches progressively adopt inefficient full checkpointing and expensive cross-node communication as context lengths grow, ultimately encountering out-of-memory failures in many cases. For example, in the 512K context length scenario with 128 GPUs, SlimPipe achieves a $1.57\times$ speedup over Megatron-LM when training Mixtral~8x7B. Meanwhile, it maintains high efficiency on other larger models, while Megatron-LM encounters OOM issues.

\noindent\textbf{Difference in Scalability.}
DeepSpeed fails to run with a 512K context length on a total of 128 GPUs (no viable configuration), because the batch size 8 is not enough for a larger DP size.
It cannot enlarge the UP size because there are only 8 query groups.
This limitation of DeepSpeed becomes severer if running on more GPUs.
When scaling to 512 GPUs, systems have to expand their DP size to saturate this amount of GPUs.
A larger DP size leads to fewer microbatches, and thus to a shorter steady phase in the pipeline.
Consequently, a larger bubble fraction hinders the scalability of Megatron-LM.
Furthermore, if the DP size continues expanding, the number of microbatches left in PP will become even less than the PP size, breaking the minimum requirements by the interleaved 1F1B scheme.
This fatal limitation prevents Megatron-LM from scaling to more GPUs.
In contrast, SlimPipe maintains quite high training efficiency with as few as 2 micorbatches in PP,
because it effectively reduces the warm-up bubbles by fine-grained slicing.

\noindent\textbf{Difference in Models.} SlimPipe's advantages become increasingly pronounced with larger-scale models. For both dense and MoE architectures, SlimPipe demonstrates superior performance as model size grows. Crucially, SlimPipe eliminates the need for either full checkpointing or costly cross-node communication during large-model training. For example, in the 512 GPUs, 512K context length scenario, SlimPipe achieves a $1.41\times$ speedup on the 8x22B model, while it only achieves a $1.04\times$ speedup on the 8x7B model. This scalability advantage suggests that SlimPipe's performance lead over alternative systems would further widen with even larger models.

In a nutshell, SlimPipe demonstrates its outstanding aptitude to train long-context LLMs at any scale.
}

\subsection{Ultra Long Context}
\input{data/extending.tex}
512K context is not the very limit of SlimpPipe.
To explore the extensibility of SlimPipe, we integrate the pipeline-parallelism-aware offloading~\cite{yuan2024accelerating} into our system.
This technique saves GPU memory by transferring portions of the activations to the host memory.
We train 4 large models with 16M tokens per iteration and
extend the context length to the maximum supported by our system on 256 GPUs.
The system configurations just follow those of Section~\ref{sec:system}, with an additional offload ratio.

The results in Table~\ref{tab:extending} show that SlimPipe maintains its high efficiency when training with spectacular context lengths. With the help of offloading, SlimPipe pushes the boundary of context length to an astonishing 4096K in training Mixtral~8x7B. It also delivers a remarkable MFU as high as 45.0\% in training Llama~70B. Now we can say with confidence that we realize the ambitious goal of training LLMs with ultra long context.

\subsection{Comparison of Pipeline Schemes}\label{sec:long_ctx_pp}
The dessert is a comparison between pipeline schemes on training LLMs with long context length.
The model is Llama~13B and the per-iteration batch size is 4.
The input sequence length increases from 32K to 512K.
8-way TP and full checkpointing are used to reduce the memory pressure.
The number of stages per device is set to 5 for both interleaved 1F1B and SlimPipe.
The number of slices is fixed to 4 for SlimPipe.
The ZB-V and V-Half schemes use \textsl{float16} in place of \textsl{bfloat16}, for the latter is not yet supported.

\input{data/long_seq_pp.tex}
\input{data/pp_mem.tex}

The MFUs are drawn as dots in Figure~\ref{fig:long_ctx_pp}, and the memory consumptions are drawn as bars in Figure~\ref{fig:pp_mem}.
The ZB-V scheme goes out of memory quite early.
It decouples the backward passes of inputs and weights and rearranges them in the pipeline schedule.
Its built-in full checkpointing implementation does not work properly in this scheme.
This flaw also restricts the maximum context length supported by the V-Half, which is designed to accumulate half the activations of the ZB-V.
These V-shaped schemes both suffer from imbalance bubbles, caused by the diverged computation time of two decoupled backward passes.
Although the default 1F1B sustains a fairly long context length of up to 256K,
its efficiency is quite low due to relatively large warm-up bubbles.
The interleaved 1F1B shows a competitive efficiency with short context lengths, but is still confined by its considerable memory overhead.
SlimPipe delivers higher efficiency than others through all context lengths,
demonstrating its outstanding ability to reduce both memory usage and pipeline bubbles.

\section{Conclusions}
This paper introduces SlimPipe, a fine-grained pipeline parallelism with uniform slicing.
Firstly, SlimPipe leverages uniform slicing to reduce GPU memory usage for activations proportionally as the PP size increases.
Additionally, this fine-grained pipeline parallelism schedule reduces bubbles during the warm-up and cool-down phases.
Furthermore, we found that uniform slicing, due to the presence of causal attention, leads to workload imbalance between different slices.
To avoid pipeline bubbles introduced by imbalanced workload, SlimPipe incorporates an attention context exchange mechanism.
Our experiments demonstrate that SlimPipe can outperform state-of-the-art training systems.

\bibliographystyle{ACM-Reference-Format}
\input{main.bbl}

\end{document}

%% file: img/mem_comparison.tex
\begin{figure}[]
    \centering
    \includegraphics[]{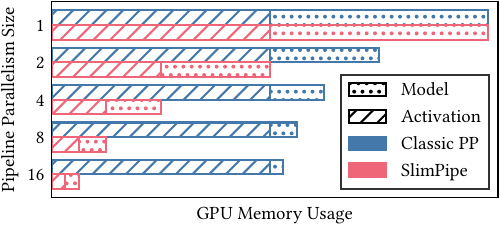}
\caption{Comparison of GPU memory footprint between Classic PP~\cite{huang2019gpipe,narayanan2021memory,narayanan2021efficient,qi2024zero,li2021terapipe} and SlimPipe across various pipeline parallelism sizes. While both approaches consume identical GPU memory for model parameters, SlimPipe's activation memory decreases proportionally as pipeline parallelism size increases, in contrast to Classic PP's constant activation memory requirements.}
\label{fig:memory_comparison}
\end{figure}

%% file: img/mem_pp_rank.tex
\begin{figure}[]
    \centering
    \begin{minipage}[t]{0.48\linewidth}
        \centering
        \includegraphics[]{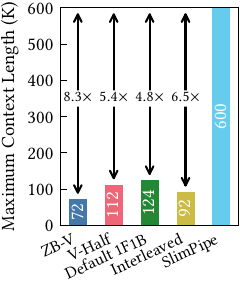}
        \caption{Maximum context lengths supported by different PP schemes in training Llama~7B with 8-way TP and 8-way PP.}
        \label{fig:max_len_comparison}
    \end{minipage}
    \hfill
    \begin{minipage}[t]{0.48\linewidth}
        \centering
        \includegraphics[]{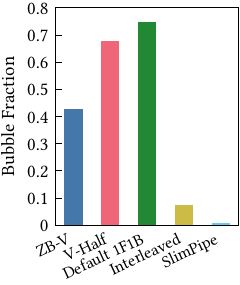}
        \caption{Theoretical bubble fractions of different PP schemes (PP size 8) in training Llama~13B with 4 microbatches and a 256K context length. }
        \label{fig:bubble_comparison}
    \end{minipage}
\end{figure}

%% file: table/compare_sched.tex
\begin{table}[ht]
    \normalfont
    \centering
    \caption{Comparison between pipeline schemes.}
    \begin{threeparttable}
    \setlength{\tabcolsep}{4pt}
    \setlength\cellspacetoplimit{2pt}
    \setlength\cellspacebottomlimit{2pt}
    \begin{tabular}{ c | Olr | Olr }
        \toprule
         Pipeline Scheme  & \multicolumn{2}{c|}{Activation Memory} & \multicolumn{2}{c}{Bubble Fraction}  \\
        \midrule
          GPipe              &   $\displaystyle \frac{m}{p}$ & $\displaystyle\uparrow\uparrow$ & $\displaystyle\frac{p-1}{m}$ & -- \\
          \hline
          TeraPipe           &   $\displaystyle \frac{m}{p}$ & $\displaystyle \uparrow\uparrow$  & $\displaystyle\frac{p - 1}{nm}$ & $\displaystyle \downarrow\downarrow$ \\
          \hline
          \makecell{PipeDream-Flush \\ (default 1F1B)}   &   $\displaystyle 1$ & \textendash & $\displaystyle \frac{p - 1}{m}$ & \textendash \\
          \hline
          Interleaved 1F1B  & $\displaystyle 1+\frac{p-1}{vp}$ & $\displaystyle \uparrow$ & $\displaystyle \frac{p - 1}{vm}$ & $\displaystyle \downarrow\downarrow$ \\
          \hline
          ZB-V   &               $\displaystyle 1$  & \textendash     & $\displaystyle \Bigl(0, \frac{2(p-1)}{3m}\Bigr)$\tnote{\dag} & $\displaystyle \downarrow$ \\
          \hline
          V-Half   &   $\displaystyle \frac{1}{2}+\frac{1}{p}$  & $\displaystyle \downarrow$     & $\displaystyle \Bigl(\frac{p}{2m}, \frac{1}{3}+\frac{p}{2m}\Bigr)$\tnote{\dag} & $\displaystyle \uparrow$ \\
          \hline
          SlimPipe  & $\displaystyle \frac{1}{p}+\frac{2(p-1)}{nvp}$ & $\displaystyle \downarrow\downarrow\downarrow$ & $\displaystyle <\frac{p-1}{nvm}$\tnote{\ddag} & $\displaystyle \downarrow\downarrow\downarrow$ \\
    \bottomrule
    \end{tabular}
    \begin{tablenotes}
      \footnotesize
      \item[\dag] It increases with longer context length and the upper bound is approximate.
      \item[\ddag] It will be $\displaystyle \frac{(p-1)p}{(n+1)nvm}$, with extremely long context length.

    \end{tablenotes}
  \end{threeparttable}
  \label{tab:sched-comparison}
  \end{table}

%% file: img/slim_1f1b.tex
\begin{figure*}[ht]
    \scriptsize
    \begin{tikzpicture}[scale=.16]
        \newcommand{\forwardheight}{3}
        \definecolor{forwardacolor}{rgb}{0.19, 0.33, 0.588}
        \definecolor{forwardbcolor}{rgb}{0.702, 0.773, 0.9}
        \definecolor{backwardacolor}{rgb}{0.663, 0.816, 0.557}
        \definecolor{backwardbcolor}{rgb}{0.216, 0.339, 0.139}
        \definecolor{offloadacolor}{rgb}{0.7, 0.35, 0}
        \definecolor{offloadbcolor}{rgb}{0.9, 0.75, 0.6}
        \definecolor{prefetchacolor}{rgb}{0.32, 0.25, 0.40}
        \definecolor{prefetchbcolor}{rgb}{0.77, 0.61, 0.88}
        \definecolor{bubbleacolor}{rgb}{0.5, 0.5, 0.5}
        \newcommand{\forwarda}[1]{%
            \draw[fill=forwardacolor] (\x,\ybase+0) rectangle node[white]{\scriptsize{\texttt{#1}}} (\x+8,\ybase+\forwardheight);
            \edef\x{\the\numexpr\x+8\relax}
        }
        \newcommand{\forwardb}[1]{%
            \draw[fill=none] (\x,\ybase+0) rectangle node[black]{\scriptsize{\texttt{#1}}} (\x+8,\ybase+\forwardheight);
            \edef\x{\the\numexpr\x+8\relax}
        }
        \newcommand{\backwarda}[1]{%
            \draw[fill=backwardacolor] (\x,\ybase+0) rectangle node[black]{\scriptsize{\texttt{#1}}} (\x+16,\ybase+\forwardheight);
            \edef\x{\the\numexpr\x+16\relax}
        }
        \newcommand{\backwardb}[1]{%
            \draw[fill=backwardacolor] (\x,\ybase+0) rectangle node[black]{\scriptsize{\texttt{#1}}} (\x+8,\ybase+\forwardheight);
            \edef\x{\the\numexpr\x+8\relax}
        }
        \newcommand{\bubblea}[1]{%
            \edef\x{\the\numexpr\x+#1*8\relax}
        }
        \newcommand{\forwardx}[2]{%
            \draw[fill=forwardacolor] (\x,\ybase+0) rectangle node[white]{\scriptsize{$\mathtt{\overset{#2}{#1}}$}} (\x+1,\ybase+\forwardheight);
            \edef\x{\the\numexpr\x+1\relax}
        }
        \newcommand{\backwardx}[2]{%
            \draw[fill=backwardacolor] (\x,\ybase+0) rectangle node[black]{$\mathtt{\overset{#2}{#1}}$} (\x+2,\ybase+\forwardheight);
            \edef\x{\the\numexpr\x+2\relax}
        }
        \newcommand{\forwardy}[2]{%
            \draw[fill=none] (\x,\ybase+0) rectangle node[black]{\scriptsize{$\mathtt{\overset{#2}{#1}}$}} (\x+1,\ybase+\forwardheight);
            \edef\x{\the\numexpr\x+1\relax}
        }
        \newcommand{\bubble}[1]{%
            \edef\x{\the\numexpr\x+#1\relax}
        }

        \def\ybase{0}
        \def\x{0}
        \draw (-4, {\ybase+\forwardheight/2}) node {\normalfont{Device 1}};
        \forwarda{1}\forwarda{2}\forwarda{3}\forwarda{4}
        \bubblea{6}\backwarda{1}

        \def\ybase{-3}
        \def\x{0}
        \draw (-4, {\ybase+\forwardheight/2}) node {\normalfont{Device 2}};
        \bubblea{1}
        \forwarda{1}\forwarda{2}\forwarda{3}
        \bubblea{4}\backwarda{1}
        \forwarda{4}\backwardb{2}

        \def\ybase{-6}
        \def\x{0}
        \draw (-4, {\ybase+\forwardheight/2}) node {\normalfont{Device 3}};
        \bubblea{2}
        \forwarda{1}\forwarda{2}
        \bubblea{2}\backwarda{1}
        \forwarda{3}\backwarda{2}
        \forwarda{4}%

        \def\ybase{-9}
        \def\x{0}
        \draw (-4, {\ybase+\forwardheight/2}) node {\normalfont{Device 4}};
        \bubblea{3}
        \forwarda{1}\backwarda{1}
        \forwarda{2}\backwarda{2}
        \forwarda{3}\backwarda{3}

        \def\ybase{-12}
        \def\x{0}
        \draw (-3, {\ybase+\forwardheight/2}) node {\normalfont{Time}};
        \draw[->] (0, {\ybase+\forwardheight/2}) -- (5, {\ybase+\forwardheight/2});

        \draw[thick] (96,4) -- (96,-10);
        \draw[thick] (97,4) -- (97,-10);

    \normalfont
    \normalsize
    \draw[ultra thick, ->] (40, -12) -- node[anchor=west] {split each input sequence into $n$ slices}(40, -17);

    \footnotesize
    \draw [decorate,decoration={brace,amplitude=5pt,mirror,raise=3ex}] (32,1) -- (0,1) node[midway,yshift=3em]{$p \cdot \frac{M_a}{p}$};
    \draw [decorate,decoration={brace,amplitude=5pt,mirror,raise=3ex}] (8,-23) -- (0,-23) node[midway,yshift=3em]{$\frac{M_a}{p}$};
    \draw [decorate,decoration={brace,amplitude=5pt,mirror,raise=3ex}] (20,-23) -- (8,-23) node[midway,yshift=3em]{$\delta \frac{M_a}{p}$};

        \def\ybase{-24}
        \def\x{0}
        \draw (-4, {\ybase+\forwardheight/2}) node {\normalfont{Device 1}};
        \forwardx{1}{1}\forwardx{1}{2}\forwardx{1}{3}\forwardx{1}{4}\forwardx{1}{5}\forwardx{1}{6}\forwardx{1}{7}\forwardx{1}{8}
        \forwardx{2}{1}\forwardx{2}{2}\forwardx{2}{3}
        \bubble{2}\forwardx{2}{4}\bubble{2}\forwardx{2}{5}\bubble{2}\forwardx{2}{6}
        \backwardx{1}{8}\forwardx{2}{7}
        \backwardx{1}{7}\forwardx{2}{8}
        \backwardx{1}{6}\forwardx{3}{1}
        \backwardx{1}{5}\forwardx{3}{2}
        \backwardx{1}{4}\forwardx{3}{3}
        \backwardx{1}{3}\forwardx{3}{4}
        \backwardx{1}{2}\forwardx{3}{5}
        \backwardx{1}{1}\forwardx{3}{6}
        \backwardx{2}{8}\forwardx{3}{7}
        \backwardx{2}{7}\forwardx{3}{8}
        \backwardx{2}{6}\forwardx{4}{1}
        \backwardx{2}{5}\forwardx{4}{2}
        \backwardx{2}{4}\forwardx{4}{3}
        \backwardx{2}{3}\forwardx{4}{4}
        \backwardx{2}{2}\forwardx{4}{5}
        \backwardx{2}{1}\forwardx{4}{6}
        \backwardx{3}{8}\forwardx{4}{7}
        \backwardx{3}{7}\forwardx{4}{8}
        \backwardx{3}{6}\forwardx{5}{1}
        \backwardx{3}{5}\forwardx{5}{2}
        \backwardx{3}{4}%

        \def\ybase{-27}
        \def\x{0}
        \draw (-4, {\ybase+\forwardheight/2}) node {\normalfont{Device 2}};
        \bubble{1}
        \forwardx{1}{1}\forwardx{1}{2}\forwardx{1}{3}\forwardx{1}{4}\forwardx{1}{5}\forwardx{1}{6}\forwardx{1}{7}\forwardx{1}{8}
        \forwardx{2}{1}\forwardx{2}{2}
        \bubble{2}\forwardx{2}{3}\bubble{2}\forwardx{2}{4}
        \backwardx{1}{8}\forwardx{2}{5}
        \backwardx{1}{7}\forwardx{2}{6}
        \backwardx{1}{6}\forwardx{2}{7}
        \backwardx{1}{5}\forwardx{2}{8}
        \backwardx{1}{4}\forwardx{3}{1}
        \backwardx{1}{3}\forwardx{3}{2}
        \backwardx{1}{2}\forwardx{3}{3}
        \backwardx{1}{1}\forwardx{3}{4}
        \backwardx{2}{8}\forwardx{3}{5}
        \backwardx{2}{7}\forwardx{3}{6}
        \backwardx{2}{6}\forwardx{3}{7}
        \backwardx{2}{5}\forwardx{3}{8}
        \backwardx{2}{4}\forwardx{4}{1}
        \backwardx{2}{3}\forwardx{4}{2}
        \backwardx{2}{2}\forwardx{4}{3}
        \backwardx{2}{1}\forwardx{4}{4}
        \backwardx{3}{8}\forwardx{4}{5}
        \backwardx{3}{7}\forwardx{4}{6}
        \backwardx{3}{6}\forwardx{4}{7}
        \backwardx{3}{5}\forwardx{4}{8}
        \backwardx{3}{4}\forwardx{5}{1}
        \backwardx{3}{3}%

        \def\ybase{-30}
        \def\x{0}
        \draw (-4, {\ybase+\forwardheight/2}) node {\normalfont{Device 3}};
        \bubble{2}
        \forwardx{1}{1}\forwardx{1}{2}\forwardx{1}{3}\forwardx{1}{4}\forwardx{1}{5}\forwardx{1}{6}\forwardx{1}{7}\forwardx{1}{8}
        \forwardx{2}{1}
        \bubble{2}\forwardx{2}{2}
        \backwardx{1}{8}\forwardx{2}{3}
        \backwardx{1}{7}\forwardx{2}{4}
        \backwardx{1}{6}\forwardx{2}{5}
        \backwardx{1}{5}\forwardx{2}{6}
        \backwardx{1}{4}\forwardx{2}{7}
        \backwardx{1}{3}\forwardx{2}{8}
        \backwardx{1}{2}\forwardx{3}{1}
        \backwardx{1}{1}\forwardx{3}{2}
        \backwardx{2}{8}\forwardx{3}{3}
        \backwardx{2}{7}\forwardx{3}{4}
        \backwardx{2}{6}\forwardx{3}{5}
        \backwardx{2}{5}\forwardx{3}{6}
        \backwardx{2}{4}\forwardx{3}{7}
        \backwardx{2}{3}\forwardx{3}{8}
        \backwardx{2}{2}\forwardx{4}{1}
        \backwardx{2}{1}\forwardx{4}{2}
        \backwardx{3}{8}\forwardx{4}{3}
        \backwardx{3}{7}\forwardx{4}{4}
        \backwardx{3}{6}\forwardx{4}{5}
        \backwardx{3}{5}\forwardx{4}{6}
        \backwardx{3}{4}\forwardx{4}{7}
        \backwardx{3}{3}\forwardx{4}{8}
        \backwardx{3}{2}%

        \def\ybase{-33}
        \def\x{0}
        \draw (-4, {\ybase+\forwardheight/2}) node {\normalfont{Device 4}};
        \bubble{3}
        \forwardx{1}{1}\forwardx{1}{2}\forwardx{1}{3}\forwardx{1}{4}\forwardx{1}{5}\forwardx{1}{6}\forwardx{1}{7}\forwardx{1}{8}
        \backwardx{1}{8}\forwardx{2}{1}
        \backwardx{1}{7}\forwardx{2}{2}
        \backwardx{1}{6}\forwardx{2}{3}
        \backwardx{1}{5}\forwardx{2}{4}
        \backwardx{1}{4}\forwardx{2}{5}
        \backwardx{1}{3}\forwardx{2}{6}
        \backwardx{1}{2}\forwardx{2}{7}
        \backwardx{1}{1}\forwardx{2}{8}
        \backwardx{2}{8}\forwardx{3}{1}
        \backwardx{2}{7}\forwardx{3}{2}
        \backwardx{2}{6}\forwardx{3}{3}
        \backwardx{2}{5}\forwardx{3}{4}
        \backwardx{2}{4}\forwardx{3}{5}
        \backwardx{2}{3}\forwardx{3}{6}
        \backwardx{2}{2}\forwardx{3}{7}
        \backwardx{2}{1}\forwardx{3}{8}
        \backwardx{3}{8}\forwardx{4}{1}
        \backwardx{3}{7}\forwardx{4}{2}
        \backwardx{3}{6}\forwardx{4}{3}
        \backwardx{3}{5}\forwardx{4}{4}
        \backwardx{3}{4}\forwardx{4}{5}
        \backwardx{3}{3}\forwardx{4}{6}
        \backwardx{3}{2}\forwardx{4}{7}
        \backwardx{3}{1}%

        \def\ybase{-36}
        \def\x{0}
        \draw (-3, {\ybase+\forwardheight/2}) node {\normalfont{Time}};
        \draw[->] (0, {\ybase+\forwardheight/2}) -- (5, {\ybase+\forwardheight/2});

        \draw[thick] (82,-20) -- (82,-34);
        \draw[thick] (83,-20) -- (83,-34);

        \def\ybase{-39}
        \def\x{6}
        \forwardx{k}{i}
        \draw(22, {\ybase+\forwardheight/2}) node {\normalfont{Forward Pass (microbatch k, slice i)}};
        \def\x{50}
        \backwardx{k}{i}
        \draw(68, {\ybase+\forwardheight/2}) node {\normalfont{Backward pass (microbatch k, slice i)}};

    \end{tikzpicture}

    \normalfont
    \caption{%
    The top figure shows the default 1F1B schedule.
    The bottom figure shows the SlimPipe schedule, where each microbatch is broken into multiple (in this case, 8) slices.
    $M_a$ represents the total activation size of one microbatch.
    $n$ and $p$ indicate the number of slices per sequence and the PP size, respectively.
    The activation memory accumulated during the warm-up phase on device 1 reduces from $M_a$ to $(1+\delta)\frac{M_a}{p}$, where $\delta=2(p-1)/n$.
    In the mean time, the warm-up bubble shrinks by about $n$ times.} \label{fig:slim_schedule}
\end{figure*}
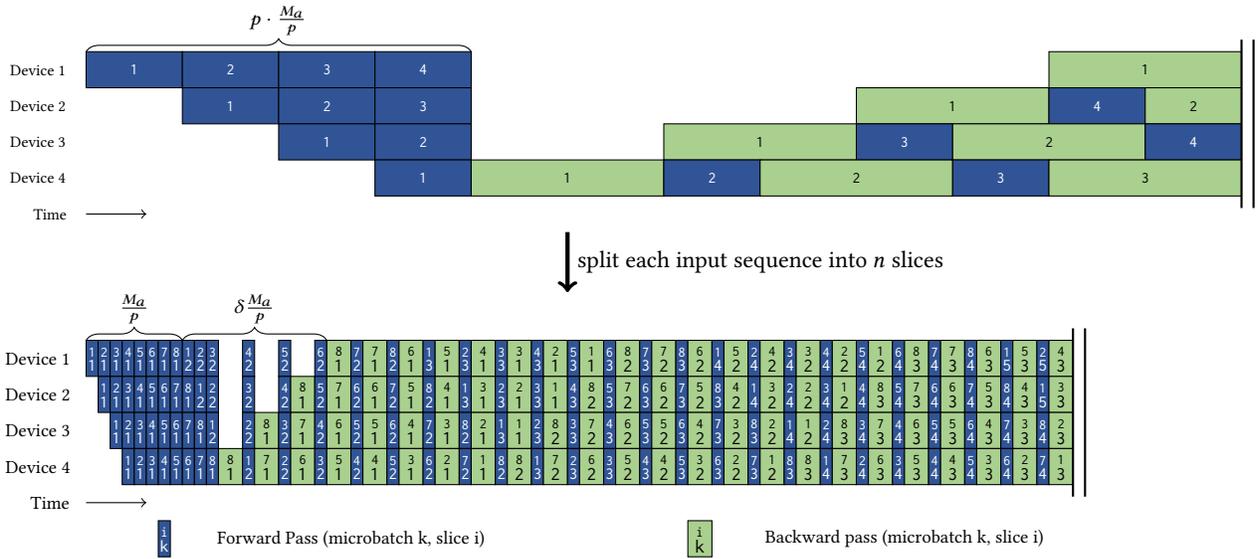

%% file: img/interleaved_slim_1f1b.tex
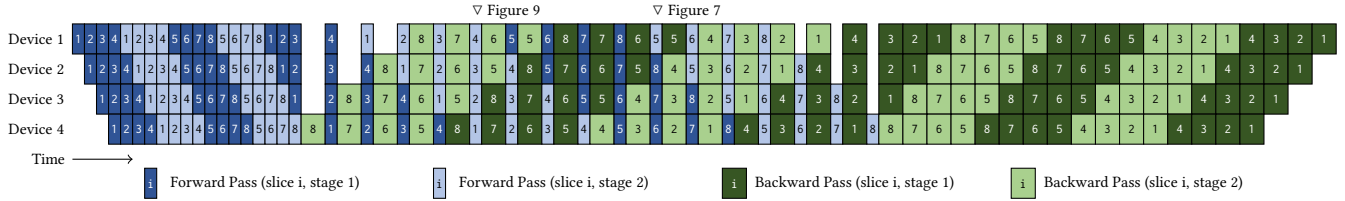
\begin{figure*}[h]
    \scriptsize
    \ttfamily
    \begin{tikzpicture}[scale=.16]
        \newcommand{\forwardheight}{2.5}
        \newcommand{\offloadheight}{1.25}
        \newcommand{\prefetchheight}{1.25}
        \definecolor{forwardacolor}{rgb}{0.19, 0.33, 0.588}
        \definecolor{forwardbcolor}{rgb}{0.702, 0.773, 0.9}
        \definecolor{backwardacolor}{rgb}{0.216, 0.339, 0.139}
        \definecolor{backwardbcolor}{rgb}{0.663, 0.816, 0.557}
        \definecolor{offloadacolor}{rgb}{0.7, 0.35, 0}
        \definecolor{offloadbcolor}{rgb}{0.9, 0.75, 0.6}
        \definecolor{prefetchacolor}{rgb}{0.32, 0.25, 0.40}
        \definecolor{prefetchbcolor}{rgb}{0.77, 0.61, 0.88}
        \definecolor{bubbleacolor}{rgb}{0.5, 0.5, 0.5}
        \newcommand{\forwarda}[1]{%
            \draw[fill=forwardacolor] (\x,\ybase+0) rectangle node[white]{\tiny{#1}} (\x+1,\ybase+\forwardheight);
            \edef\x{\the\numexpr\x+1\relax}
        }
        \newcommand{\forwardb}[1]{%
            \draw[fill=forwardbcolor] (\x,\ybase+0) rectangle node[black]{\tiny{#1}} (\x+1,\ybase+\forwardheight);
            \edef\x{\the\numexpr\x+1\relax}
        }
        \newcommand{\backwarda}[1]{%
            \draw[fill=backwardacolor] (\x,\ybase+0) rectangle node[white]{\tiny{#1}} (\x+2,\ybase+\forwardheight);
            \edef\x{\the\numexpr\x+2\relax}
        }
        \newcommand{\backwardb}[1]{%
            \draw[fill=backwardbcolor] (\x,\ybase+0) rectangle node[black]{\tiny{#1}} (\x+2,\ybase+\forwardheight);
            \edef\x{\the\numexpr\x+2\relax}
        }
        \newcommand{\bubblea}[1]{%
            \edef\x{\the\numexpr\x+#1\relax}
        }

        \draw (36,3.5) node {\normalfont{$\bigtriangledown$ Figure~\ref{fig:post_bubble}}};
        \draw (51,3.5) node {\normalfont{$\bigtriangledown$ Figure~\ref{fig:imba_bubble}}};

        \def\ybase{0}
        \def\x{0}
        \draw (-3, {\ybase+\forwardheight/2}) node {\normalfont{Device 1}};
        \forwarda{1}\forwarda{2}\forwarda{3}\forwarda{4}
        \forwardb{1}\forwardb{2}\forwardb{3}\forwardb{4}
        \forwarda{5}\forwarda{6}\forwarda{7}\forwarda{8}
        \forwardb{5}\forwardb{6}\forwardb{7}\forwardb{8}
        \forwarda{1}\forwarda{2}
        \forwarda{3}\bubblea{2}
        \forwarda{4}\bubblea{2}
        \forwardb{1}\bubblea{2}
        \forwardb{2}\backwardb{8}
        \forwardb{3}\backwardb{7}
        \forwardb{4}\backwardb{6}
        \forwarda{5}\backwardb{5}
        \forwarda{6}\backwarda{8}
        \forwarda{7}\backwarda{7}
        \forwarda{8}\backwarda{6}
        \forwardb{5}\backwarda{5}
        \forwardb{6}\backwardb{4}
        \forwardb{7}\backwardb{3}
        \forwardb{8}\backwardb{2}
        \bubblea{1}\backwardb{1}
        \bubblea{1}\backwarda{4}
        \bubblea{1}\backwarda{3}
        \backwarda{2}\backwarda{1}
        \backwardb{8}\backwardb{7}\backwardb{6}\backwardb{5}
        \backwarda{8}\backwarda{7}\backwarda{6}\backwarda{5}
        \backwardb{4}\backwardb{3}\backwardb{2}\backwardb{1}
        \backwarda{4}\backwarda{3}\backwarda{2}\backwarda{1}

        \def\ybase{-2.5}
        \def\x{0}
        \draw (-3, {\ybase+\forwardheight/2}) node {\normalfont{Device 2}};
        \bubblea{1}
        \forwarda{1}\forwarda{2}\forwarda{3}\forwarda{4}
        \forwardb{1}\forwardb{2}\forwardb{3}\forwardb{4}
        \forwarda{5}\forwarda{6}\forwarda{7}\forwarda{8}
        \forwardb{5}\forwardb{6}\forwardb{7}\forwardb{8}
        \forwarda{1}\forwarda{2}\bubblea{2}
        \forwarda{3}\bubblea{2}
        \forwarda{4}\backwardb{8}
        \forwardb{1}\backwardb{7}
        \forwardb{2}\backwardb{6}
        \forwardb{3}\backwardb{5}
        \forwardb{4}\backwarda{8}
        \forwarda{5}\backwarda{7}
        \forwarda{6}\backwarda{6}
        \forwarda{7}\backwarda{5}
        \forwarda{8}\backwardb{4}
        \forwardb{5}\backwardb{3}
        \forwardb{6}\backwardb{2}
        \forwardb{7}\backwardb{1}
        \forwardb{8}\backwarda{4}
        \bubblea{1}\backwarda{3}
        \bubblea{1}\backwarda{2}\backwarda{1}
        \backwardb{8}\backwardb{7}\backwardb{6}\backwardb{5}
        \backwarda{8}\backwarda{7}\backwarda{6}\backwarda{5}
        \backwardb{4}\backwardb{3}\backwardb{2}\backwardb{1}
        \backwarda{4}\backwarda{3}\backwarda{2}\backwarda{1}

        \def\ybase{-5}
        \def\x{0}
        \draw (-3, {\ybase+\forwardheight/2}) node {\normalfont{Device 3}};
        \bubblea{2}
        \forwarda{1}\forwarda{2}\forwarda{3}\forwarda{4}
        \forwardb{1}\forwardb{2}\forwardb{3}\forwardb{4}
        \forwarda{5}\forwarda{6}\forwarda{7}\forwarda{8}
        \forwardb{5}\forwardb{6}\forwardb{7}\forwardb{8}
        \forwarda{1}\bubblea{2}
        \forwarda{2}\backwardb{8}
        \forwarda{3}\backwardb{7}
        \forwarda{4}\backwardb{6}
        \forwardb{1}\backwardb{5}
        \forwardb{2}\backwarda{8}
        \forwardb{3}\backwarda{7}
        \forwardb{4}\backwarda{6}
        \forwarda{5}\backwarda{5}
        \forwarda{6}\backwardb{4}
        \forwarda{7}\backwardb{3}
        \forwarda{8}\backwardb{2}
        \forwardb{5}\backwardb{1}
        \forwardb{6}\backwarda{4}
        \forwardb{7}\backwarda{3}
        \forwardb{8}\backwarda{2}
        \bubblea{1}\backwarda{1}
        \backwardb{8}\backwardb{7}\backwardb{6}\backwardb{5}
        \backwarda{8}\backwarda{7}\backwarda{6}\backwarda{5}
        \backwardb{4}\backwardb{3}\backwardb{2}\backwardb{1}
        \backwarda{4}\backwarda{3}\backwarda{2}\backwarda{1}

        \def\ybase{-7.5}
        \def\x{0}
        \draw (-3, {\ybase+\forwardheight/2}) node {\normalfont{Device 4}};
        \bubblea{3}
        \forwarda{1}\forwarda{2}\forwarda{3}\forwarda{4}
        \forwardb{1}\forwardb{2}\forwardb{3}\forwardb{4}
        \forwarda{5}\forwarda{6}\forwarda{7}\forwarda{8}
        \forwardb{5}\forwardb{6}\forwardb{7}\forwardb{8}
        \backwardb{8}\forwarda{1}
        \backwardb{7}\forwarda{2}
        \backwardb{6}\forwarda{3}
        \backwardb{5}\forwarda{4}
        \backwarda{8}\forwardb{1}
        \backwarda{7}\forwardb{2}
        \backwarda{6}\forwardb{3}
        \backwarda{5}\forwardb{4}
        \backwardb{4}\forwarda{5}
        \backwardb{3}\forwarda{6}
        \backwardb{2}\forwarda{7}
        \backwardb{1}\forwarda{8}
        \backwarda{4}\forwardb{5}
        \backwarda{3}\forwardb{6}
        \backwarda{2}\forwardb{7}
        \backwarda{1}\forwardb{8}
        \backwardb{8}\backwardb{7}\backwardb{6}\backwardb{5}
        \backwarda{8}\backwarda{7}\backwarda{6}\backwarda{5}
        \backwardb{4}\backwardb{3}\backwardb{2}\backwardb{1}
        \backwarda{4}\backwarda{3}\backwarda{2}\backwarda{1}

        \def\ybase{-10}
        \def\x{0}
        \draw (-2, {\ybase+\forwardheight/2}) node {\normalfont{Time}};
        \draw[->] (0, {\ybase+\forwardheight/2}) -- (5, {\ybase+\forwardheight/2});

        \def\ybase{-12}
        \def\x{6}
        \forwarda{i}
        \draw(16, {\ybase+\forwardheight/2}) node {\normalfont{Forward Pass (slice i, stage 1)}};
        \def\x{30}
        \forwardb{i}
        \draw(40, {\ybase+\forwardheight/2}) node {\normalfont{Forward Pass (slice i, stage 2)}};
        \def\x{54}
        \backwarda{i}
        \draw(65, {\ybase+\forwardheight/2}) node {\normalfont{Backward Pass (slice i, stage 1)}};
        \def\x{78}
        \backwardb{i}
        \draw(89, {\ybase+\forwardheight/2}) node {\normalfont{Backward Pass (slice i, stage 2)}};

    \end{tikzpicture}
    \normalfont
    \caption{SlimPipe in its interleaving form.
    Each device is assigned 2 stages. Dark colors show the first stage and light colors show the second stage.
    For simplicity, the microbatch number is not labeled and the number in the box is the slice number.
    Note that there are only two microbatches in this pipeline, each split into 8 slices.
    Whereas in the classic interleaved 1F1B pipeline, at least 4 (the PP size) microbatches are required.
    More details are depicted in Figure~\ref{fig:imba_bubble} and Figure~\ref{fig:post_bubble}.}
    \label{fig:interleaved_slim_schedule}
\end{figure*}

%% file: img/memory_red.tex
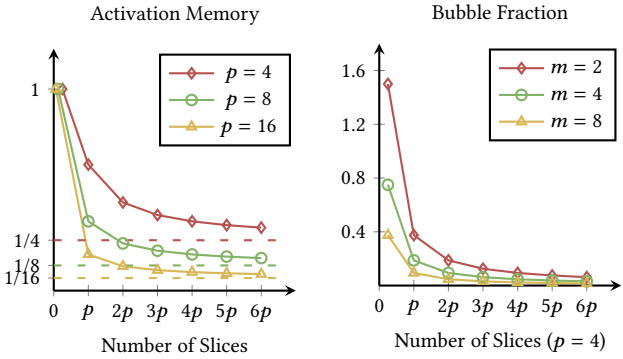
\begin{figure}[h]
    \small
    \centering
    \begin{subfigure}[t]{0.48\linewidth}
    \centering
    \begin{tikzpicture}
      \definecolor{attncolora}{RGB}{184 84 80}
      \definecolor{attncolorb}{RGB}{130 179 102}
      \definecolor{attncolorc}{RGB}{214 182 86}
      \definecolor{attncolord}{RGB}{108 142 191}
      \definecolor{attncolore}{RGB}{225, 213, 231}
      \definecolor{attncolorf}{RGB}{255, 230, 204}
    \begin{axis}[
        axis lines = left,
        width=4.8cm, height=4.8cm,
        scaled y ticks=false,
        xtick={0, 1, 2, 3, 4, 5, 6},
        ytick={6.25, 12.5, 25, 100},
        xlabel={Number of Slices},
        title={Activation Memory},
        ylabel style={yshift=-10pt},
        xticklabel style={},
        yticklabel style={},
        xticklabels={$0$, $p$, $2p$, $3p$, $4p$, $5p$, $6p$}, scaled x ticks=false,
        yticklabels={1/16, 1/8, 1/4, 1}, scaled y ticks=false,
        xmin=0, xmax=7,
        ymin=0, ymax=120,
        thick,
        legend style={
        at={(0.98,0.98)},
        anchor=north east, legend columns=1}
    ]
    \addplot[attncolora, mark=diamond] coordinates { %
        (0.25, 100) (1, 62.5) (2, 43.75) (3, 37.5) (4, 34.375) (5, 32.50) (6, 31.25)
        };
    \addplot[attncolorb, mark=o] coordinates { %
        (0.125, 100) (1, 34.375) (2, 23.4375) (3, 19.792) (4, 17.969) (5, 16.875) (6, 16.146)
    };
    \addplot[attncolorc, mark=triangle] coordinates { %
        (0.0625, 100) (1, 17.969) (2, 12.109) (3, 10.156) (4, 9.180) (5, 8.594) (6, 8.203)
    };
    \addlegendentry{$p=4$}
    \addlegendentry{$p=8$}
    \addlegendentry{$p=16$}
    \addplot[color=attncolora, loosely dashed, domain=0:6.5] {25};
    \addplot[color=attncolorb, loosely dashed, domain=0:6.5] {12.5};
    \addplot[color=attncolorc, loosely dashed, domain=0:6.5] {6.25};

    \end{axis}
    \end{tikzpicture}
    \caption{Activation memory reduced by SlimPipe across different PP sizes.}
    \label{fig:sub1}
    \end{subfigure}
    \hfill
    \begin{subfigure}[t]{0.48\linewidth}
    \centering
    \begin{tikzpicture}
      \definecolor{attncolora}{RGB}{184 84 80}
      \definecolor{attncolorb}{RGB}{130 179 102}
      \definecolor{attncolorc}{RGB}{214 182 86}
      \definecolor{attncolord}{RGB}{108 142 191}
      \definecolor{attncolore}{RGB}{225, 213, 231}
      \definecolor{attncolorf}{RGB}{255, 230, 204}
    \begin{axis}[
        axis lines = left,
        width=4.8cm, height=4.8cm,
        xtick={0, 1, 2, 3, 4, 5, 6},
        ytick={0.4, 0.8, 1.2, 1.6},
        xlabel={Number of Slices ($p=4$)},
        title={Bubble Fraction},
        xticklabel style={},
        yticklabel style={},
        xticklabels={$0$, $p$, $2p$, $3p$, $4p$, $5p$, $6p$}, scaled x ticks=false,
        yticklabels={0.4, 0.8, 1.2, 1.6},  scaled y ticks=false,
        xmin=0, xmax=7,
        ymin=0, ymax=1.8,
        thick,
        legend style={
        at={(0.98,0.98)},
        anchor=north east, legend columns=1}
    ]
    \addplot[attncolora, mark=diamond] coordinates {
        (0.25, 1.5) (1, 0.375) (2, 0.1875) (3, 0.125) (4, 0.09375) (5, 0.075) (6, 0.0625)
    };
    \addplot[attncolorb, mark=o] coordinates {
        (0.25, 0.75) (1, 0.1875) (2, 0.09375) (3, 0.0625) (4, 0.046875) (5, 0.0375) (6, 0.03125)
    };
    \addplot[attncolorc, mark=triangle] coordinates {
        (0.25, 0.375) (1, 0.09375) (2, 0.046875) (3, 0.03125) (4, 0.0234375) (5, 0.01875) (6, 0.015625)
    };
    \addlegendentry{$m=2$}
    \addlegendentry{$m=4$}
    \addlegendentry{$m=8$}

    \end{axis}
    \end{tikzpicture}
    \caption{Bubble fraction reduced by SlimPipe across different microbatch numbers. The PP size is fixed to 4.}
    \label{fig:sub2}
    \end{subfigure}
    \caption{SlimPipe reduces both activation memory and bubble fraction.
        (a) The activation memories are equally large without uniform slicing (default 1F1B),
            and are decreasing from $1$ towards $1/p$ separately as the number of slices increases.
        (b) The bubble fraction are quite large without uniform slicing (default 1F1B),
            and are decreasing to near-zero as the number of slices increases.}
    \label{fig:memory_dec}
\end{figure}

%% file: img/imba.tex
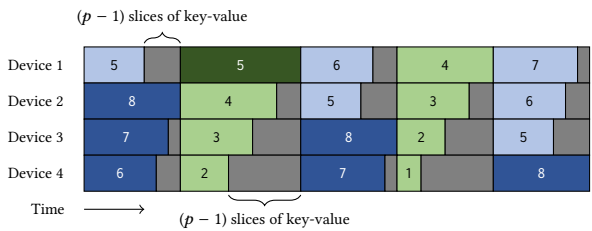
\begin{figure}[h]
    \scriptsize
    \ttfamily
    \begin{tikzpicture}[scale=.16]
        \newcommand{\forwardheight}{3}
        \definecolor{forwardacolor}{rgb}{0.19, 0.33, 0.588}
        \definecolor{forwardbcolor}{rgb}{0.702, 0.773, 0.9}
        \definecolor{backwardacolor}{rgb}{0.216, 0.339, 0.139}
        \definecolor{backwardbcolor}{rgb}{0.663, 0.816, 0.557}
        \definecolor{offloadacolor}{rgb}{0.7, 0.35, 0}
        \definecolor{offloadbcolor}{rgb}{0.9, 0.75, 0.6}
        \definecolor{prefetchacolor}{rgb}{0.32, 0.25, 0.40}
        \definecolor{prefetchbcolor}{rgb}{0.77, 0.61, 0.88}
        \definecolor{bubbleacolor}{rgb}{0.5, 0.5, 0.5}
        \newcommand{\forwarda}[1]{%
            \draw[fill=forwardacolor] (\x,\ybase+0) rectangle node[white]{#1} (\x+#1,\ybase+\forwardheight);
            \edef\x{\the\numexpr\x+#1\relax}
        }
        \newcommand{\forwardb}[1]{%
            \draw[fill=forwardbcolor] (\x,\ybase+0) rectangle node[black]{#1} (\x+#1,\ybase+\forwardheight);
            \edef\x{\the\numexpr\x+#1\relax}
        }
        \newcommand{\backwarda}[1]{%
            \draw[fill=backwardacolor] (\x,\ybase+0) rectangle node[white]{#1} (\x+2*#1,\ybase+\forwardheight);
            \edef\x{\the\numexpr\x+2*#1\relax}
        }
        \newcommand{\backwardb}[1]{%
            \draw[fill=backwardbcolor] (\x,\ybase+0) rectangle node[black]{#1} (\x+2*#1,\ybase+\forwardheight);
            \edef\x{\the\numexpr\x+2*#1\relax}
        }
        \newcommand{\bubblea}[1]{%
            \draw[fill=bubbleacolor] (\x,\ybase+0) rectangle (\x+#1,\ybase+\forwardheight);
            \edef\x{\the\numexpr\x+#1\relax}
        }

        \def\ybase{0}
        \def\x{0}
        \draw (-4, {\ybase+\forwardheight/2}) node {\normalfont{Device 1}};
        \forwardb{5}\bubblea{3}
        \backwarda{5}
        \forwardb{6}\bubblea{2}
        \backwardb{4}
        \forwardb{7}\bubblea{1}

        \def\ybase{-3}
        \def\x{0}
        \draw (-4, {\ybase+\forwardheight/2}) node {\normalfont{Device 2}};
        \forwarda{8}
        \backwardb{4}\bubblea{2}
        \forwardb{5}\bubblea{3}
        \backwardb{3}\bubblea{2}
        \forwardb{6}\bubblea{2}

        \def\ybase{-6}
        \def\x{0}
        \draw (-4, {\ybase+\forwardheight/2}) node {\normalfont{Device 3}};
        \forwarda{7}\bubblea{1}
        \backwardb{3}\bubblea{4}
        \forwarda{8}
        \backwardb{2}\bubblea{4}
        \forwardb{5}\bubblea{3}

        \def\ybase{-9}
        \def\x{0}
        \draw (-4, {\ybase+\forwardheight/2}) node {\normalfont{Device 4}};
        \forwarda{6}\bubblea{2}
        \backwardb{2}\bubblea{6}
        \forwarda{7}\bubblea{1}
        \backwardb{1}\bubblea{6}
        \forwarda{8}

         \def\ybase{-12}
         \def\x{0}
         \draw (-3, {\ybase+\forwardheight/2}) node {\normalfont{Time}};
         \draw[->] (0, {\ybase+\forwardheight/2}) -- (5, {\ybase+\forwardheight/2});

         \draw [decorate,decoration={brace,amplitude=4pt,mirror,raise=3ex}] (8,1.5) -- (5,1.5) node[midway,yshift=3em]{\normalfont{($p-1$) slices of key-value}};
         \draw [decorate,decoration={brace,amplitude=4pt,mirror,raise=3ex}] (12,-7.5) -- (18,-7.5) node[midway,yshift=-3em]{\normalfont{($p-1$) slices of key-value}};

    \end{tikzpicture}

    \normalfont
    \caption{A closer look at the SlimPipe timeline with imbalance bubble exhibited.
    The beginnings of passes are aligned across devices to clearly reveal the bubbles' shape.
    The actual timeline will be more elaborated with propagated bubbles.
    } \label{fig:imba_bubble}
\end{figure}

%% file: img/attn_balance_2.tex
\begin{figure}[h]
    \scriptsize
    \ttfamily

    \begin{tikzpicture}[scale=.32]
        \newcommand{\attnheight}{1.5}
        \definecolor{attncolora}{RGB}{218, 232, 252}
        \definecolor{attncolorb}{RGB}{255, 242, 204}
        \definecolor{attncolorc}{RGB}{213, 232, 212}
        \definecolor{attncolord}{RGB}{248, 206, 204}
        \definecolor{attncolore}{RGB}{225, 213, 231}
        \definecolor{attncolorf}{rgb}{0.9, 0.75, 0.6} %
        \definecolor{bubbleacolor}{rgb}{0.5, 0.5, 0.5}
        \newcommand{\attn}[2]{%
            \draw[fill=#2] (\x,\ybase) rectangle node[black]{$\mathtt{#1}$} (\x+3,\ybase+\attnheight);
            \edef\x{\the\numexpr\x+3\relax}
        }
        \newcommand{\attnr}[2]{%
            \draw[densely dashed, fill=#2] (\x,\ybase) rectangle node[black]{$\mathtt{#1}$} (\x+3,\ybase+\attnheight);
            \edef\x{\the\numexpr\x+3\relax}
        }
        \newcommand{\bubblea}[1]{%
            \edef\x{\the\numexpr\x+#1\relax}
        }

        \def\ybase{-16}
        \def\x{0}
        \draw (-2, {\ybase+\attnheight/2}) node {\normalfont{Device 1}};
        \def\x{0}
        \attn{Q_7, K_1, V_1}{attncolora}
        \attn{Q_7, K_2, V_2}{attncolora}
        \attn{Q_7, K_3, V_3}{attncolora}
        \attn{Q_7, K_4, V_4}{attncolora}
        \attn{Q_7, K_5, V_5}{attncolora}
        \attnr{Q_7, K_6, V_6}{none}
        \attnr{Q_7, K_7, V_7}{none}

        \def\ybase{-18}
        \def\x{0}
        \draw (-2, {\ybase+\attnheight/2}) node {\normalfont{Device 2}};
        \def\x{0}
        \attn{Q_6, K_1, V_1}{attncolorb}
        \attn{Q_6, K_2, V_2}{attncolorb}
        \attn{Q_6, K_3, V_3}{attncolorb}
        \attn{Q_6, K_4, V_4}{attncolorb}
        \attn{Q_6, K_5, V_5}{attncolorb}
        \attnr{Q_6, K_6, V_6}{none}

        \def\ybase{-20}
        \def\x{0}
        \draw (-2, {\ybase+\attnheight/2}) node {\normalfont{Device 3}};
        \def\x{0}
        \attn{Q_5, K_1, V_1}{attncolorc}
        \attn{Q_5, K_2, V_2}{attncolorc}
        \attn{Q_5, K_3, V_3}{attncolorc}
        \attn{Q_5, K_4, V_4}{attncolorc}
        \attn{Q_5, K_5, V_5}{attncolorc}

        \def\ybase{-22}
        \def\x{0}
        \draw (-2, {\ybase+\attnheight/2}) node {\normalfont{Device 4}};
        \def\x{0}
        \attn{Q_4, K_1, V_1}{attncolord}
        \attn{Q_4, K_2, V_2}{attncolord}
        \attn{Q_4, K_3, V_3}{attncolord}
        \attn{Q_4, K_4, V_4}{attncolord}

        \def\ybase{-24}
        \def\x{0}
        \draw (-2, {\ybase+\attnheight/2}) node {\normalfont{Device 5}};
        \def\x{0}
        \attn{Q_3, K_1, V_1}{attncolore}
        \attn{Q_3, K_2, V_2}{attncolore}
        \attn{Q_3, K_3, V_3}{attncolore}
        \attnr{Q_6, K_6, V_6}{attncolorb}

        \def\ybase{-26}
        \def\x{0}
        \draw (-2, {\ybase+\attnheight/2}) node {\normalfont{Device 6}};
        \def\x{0}
        \attn{Q_2, K_1, V_1}{attncolorf}
        \attn{Q_2, K_2, V_2}{attncolorf}
        \attnr{Q_7, K_6, V_6}{attncolora}
        \attnr{Q_7, K_7, V_7}{attncolora}

    \end{tikzpicture}

    \normalfont
    \caption{The attention workloads are rebalanced by exchanging the context among pipeline devices.
    The local query and portions of the local key-value are sent to another device and the partial attention output will be sent back after the calculation being completed there.}
    \label{fig:attn_balance}
\end{figure}
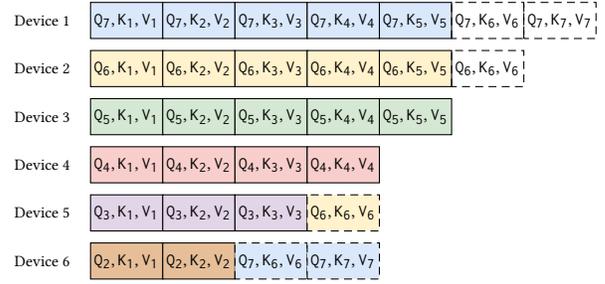

%% file: img/post_bubble.tex
\begin{figure}[h]
    \scriptsize
    \ttfamily
    \begin{tikzpicture}[scale=.16]
        \newcommand{\forwardheight}{3}
        \definecolor{forwardacolor}{rgb}{0.19, 0.33, 0.588}
        \definecolor{forwardbcolor}{rgb}{0.702, 0.773, 0.9}
        \definecolor{backwardacolor}{rgb}{0.216, 0.339, 0.139}
        \definecolor{backwardbcolor}{rgb}{0.663, 0.816, 0.557}
        \definecolor{offloadacolor}{rgb}{0.7, 0.35, 0}
        \definecolor{offloadbcolor}{rgb}{0.9, 0.75, 0.6}
        \definecolor{prefetchacolor}{rgb}{0.32, 0.25, 0.40}
        \definecolor{prefetchbcolor}{rgb}{0.77, 0.61, 0.88}
        \definecolor{bubbleacolor}{rgb}{0.5, 0.5, 0.5}
        \newcommand{\forwarda}[1]{%
            \draw[fill=forwardacolor] (\x,\ybase+0) rectangle node[white]{#1} (\x+4,\ybase+\forwardheight);
            \edef\x{\the\numexpr\x+4\relax}
        }
        \newcommand{\forwardb}[1]{%
            \draw[fill=forwardbcolor] (\x,\ybase+0) rectangle node[black]{#1} (\x+4,\ybase+\forwardheight);
            \edef\x{\the\numexpr\x+4\relax}
        }
        \newcommand{\backwarda}[1]{%
            \draw[fill=backwardacolor] (\x,\ybase+0) rectangle node[white]{#1} (\x+8,\ybase+\forwardheight);
            \edef\x{\the\numexpr\x+8\relax}
        }
        \newcommand{\backwardb}[1]{%
            \draw[fill=backwardbcolor] (\x,\ybase+0) rectangle node[black]{#1} (\x+8,\ybase+\forwardheight);
            \edef\x{\the\numexpr\x+8\relax}
        }
        \newcommand{\postforwarda}[1]{%
            \draw[fill=offloadbcolor] (\x,\ybase+0) rectangle node[black]{#1} (\x+6,\ybase+\forwardheight);
            \edef\x{\the\numexpr\x+6\relax}
        }
        \newcommand{\postforwardb}[1]{%
            \draw [dotted, fill=offloadbcolor] (\x,\ybase+0) rectangle node[black]{#1} (\x+2,\ybase+\forwardheight);
            \edef\x{\the\numexpr\x+2\relax}
        }
        \newcommand{\bubblea}[1]{%
            \draw[fill=bubbleacolor] (\x,\ybase+0) rectangle (\x+#1,\ybase+\forwardheight);
            \edef\x{\the\numexpr\x+#1\relax}
        }

        \def\ybase{0}
        \def\x{0}
        \draw (-4, {\ybase+\forwardheight/2}) node {\normalfont{Device 1}};
        \forwardb{4}\bubblea{6}\backwardb{6}

        \def\x{28}
        \draw (-4, {\ybase+\forwardheight/2}) node {\normalfont{Device 1}};
        \forwardb{4}\postforwardb{G}\backwardb{6}

        \def\ybase{-3}
        \def\x{0}
        \draw (-4, {\ybase+\forwardheight/2}) node {\normalfont{Device 2}};
        \forwardb{3}\bubblea{6}\backwardb{5}

        \def\x{28}
        \draw (-4, {\ybase+\forwardheight/2}) node {\normalfont{Device 2}};
        \forwardb{3}\postforwardb{E}\backwardb{5}

        \def\ybase{-6}
        \def\x{0}
        \draw (-4, {\ybase+\forwardheight/2}) node {\normalfont{Device 3}};
        \forwardb{2}\bubblea{6}\backwarda{8}

        \def\x{28}
        \draw (-4, {\ybase+\forwardheight/2}) node {\normalfont{Device 3}};
        \forwardb{2}\postforwardb{M}\backwarda{8}

        \def\ybase{-9}
        \def\x{0}
        \draw (-4, {\ybase+\forwardheight/2}) node {\normalfont{Device 4}};
        \forwardb{1}\postforwarda{GEMM}\backwarda{7}

        \def\x{28}
        \draw (-4, {\ybase+\forwardheight/2}) node {\normalfont{Device 4}};
        \forwardb{1}\postforwardb{M}\backwarda{7}

        \draw (32, 3) rectangle (34, -9); %
        \draw[ultra thick, ->] (20, -3) -- (25, -3);

        \def\ybase{-12}
        \def\x{0}
        \draw (-3, {\ybase+\forwardheight/2}) node {\normalfont{Time}};
        \draw[->] (0, {\ybase+\forwardheight/2}) -- (5, {\ybase+\forwardheight/2});
        \draw (25, {\ybase+\forwardheight/2}) node {\normalfont{Time}};
        \draw[->] (28, {\ybase+\forwardheight/2}) -- (33, {\ybase+\forwardheight/2});

    \end{tikzpicture}

    \normalfont
    \caption{A SlimPipe timeline segment containing the output layer (GEMM).
    The left part shows the GEMM just assigned to the last device.
    The right part shows the GEMM distributed to all devices,
    where the bubble vanishes.
    } \label{fig:post_bubble}
\end{figure}
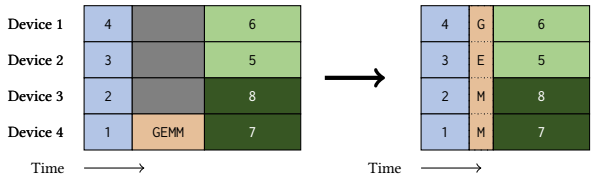

%% file: table/model_config.tex
\begin{table}[h]
    \centering
    \caption{Models used in evaluation.}
    \begin{threeparttable}
    \small
    \begin{tabular}{c | l c | l }
      $a$ & number of attention heads & $h$ & hidden dimension size \\
      $g$ & number of query groups    & $H$ & FFN dimension size \\
    \end{tabular}
    \setlength{\tabcolsep}{5pt}
    \normalsize
    \begin{tabular}{c | c | c | c | c | c | c }
      \toprule
      Model           & $L$ & $a$ & $g$ & $h$ & $H$   & \#Params.\tnote{\dag} \\
      \midrule
      Llama 13B       & 40   & 40 & --  & 5120  & 13824  & $13.3\times10^9$  \\
      Llama 70B       & 80   & 64 & 8   & 8192  & 28672  & $69.5\times10^9$   \\
      Llama 149B      & 96   & 96 & 8   & 12288 & 32768  & $148.9\times10^9$   \\
      Mixtral 8x7B  & 32 & 32 & 8    & 4096  & 14336  & $47.0\times10^9$  \\
      Mixtral 8x22B & 56 & 48 & 8    & 6144  & 16384 & $141.0\times10^9$ \\
      \bottomrule
    \end{tabular}
    \begin{tablenotes}
      \footnotesize
      \item[\dag] Including parameters in the \num{128000} sized vocabulary.
    \end{tablenotes}
  \end{threeparttable}
\label{tab:model_config}
\end{table}

%% file: data/memory_by_p.tex
\begin{figure}[h]
    \centering
    \includegraphics[]{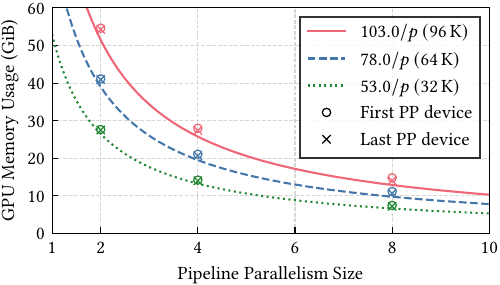}

\caption{The memory reduced by the PP size $p$. The measured values align well with our theoretical model.}
\label{fig:mem_by_p}
\end{figure}

%% file: data/tps_vs_n.tex
\begin{figure}[h]
    \centering
    \includegraphics[]{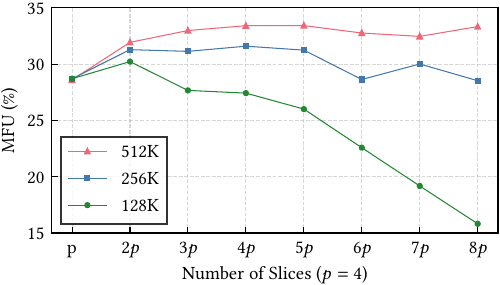}

    \caption{MFU of training the Llama~13B model with different context lengths.
            The PP size is 4 and the numbers of slices increases from $p$ to $8p$.
    }
    \label{fig:tgs-comparison}
    \end{figure}

%% file: img/system.tex
\begin{figure*}[t]
    \centering
    \includegraphics[]{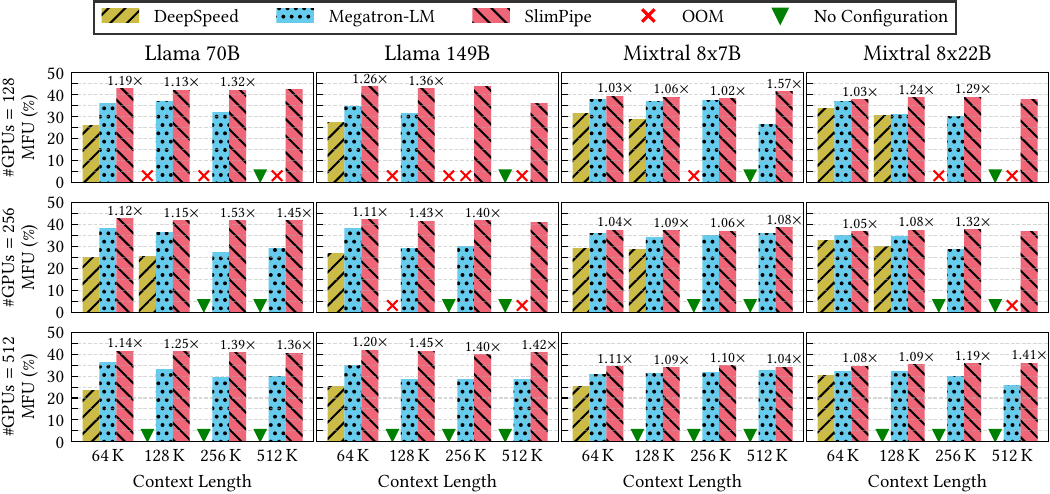}
\caption{System performance comparison between DeepSpeed, Megatron-LM, and SlimPipe across different models and GPU counts. The annotations above the bars for SlimPipe represent its relative speedup over Megatron-LM. A green triangle marker indicates that no viable configuration could be found under this condition, and a red cross indicates that all candidate configurations results in an Out-Of-Memory (OOM) error.}
\label{fig:system}
\end{figure*}

%% file: data/extending.tex
\begin{table}[h]
    \centering
    \caption{System performance of training 4 LLMs with the maximum supported context length, by using at most 256 GPUs. The selective checkpointing is enabled uniformly while the offloading ratio is adaptive.}
    \begin{threeparttable}
    \setlength{\tabcolsep}{3pt}
    \begin{tabular}{c | c | ccccccc | c }
    \toprule
    Model & Context & $t$ & $c$ & $e$ & $d$ & $p$ & $n$ & offload & MFU \\
    \midrule
    Llama~70B & 2048K & 4 & 4 & \textendash & 1 & 16 & $4p$ & 75\% &  45.0\% \\
    Llama~149B & 1024K & 4 & 2 & \textendash & 1 & 32 & $2p$ & 80\% &  43.7\% \\
    Mixtral~8x7B & 4096K & 1 & 16 & 8 & 1 & 16 & $4p$ & 95\% & 40.0\% \\
    Mixtral~8x22B\tnote{\dag}\, & 2048K & 1 & 8 & 8 & 1 & 28 & $4p$ & 100\% &  42.0\% \\
    \bottomrule
    \end{tabular}
    \begin{tablenotes}
      \footnotesize
      \item[\dag] 224 GPUs are actually used because of the PP size 28.
    \end{tablenotes}
  \end{threeparttable}
  \label{tab:extending}
  \end{table}

%% file: data/long_seq_pp.tex
\begin{figure}[h]
  \centering
  \includegraphics[]{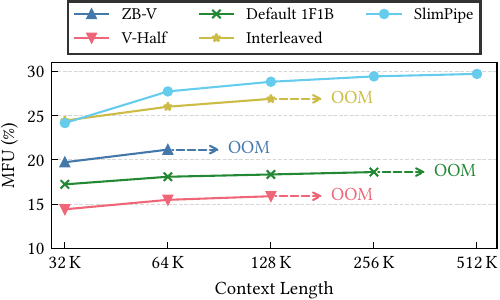}

  \caption{Comparison of MFU across different PP schemes.}
  \label{fig:long_ctx_pp}
\end{figure}

%% file: data/pp_mem.tex
\begin{figure}[h]
  \centering
  \includegraphics[]{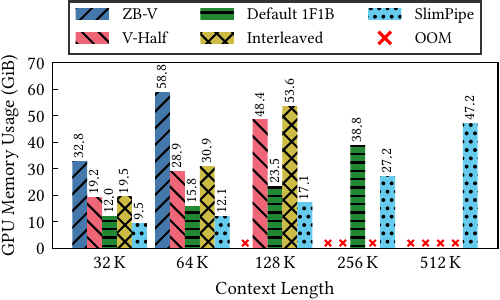}
  \caption{Comparison of GPU memory usage across different PP schemes.}
  \label{fig:pp_mem}
\end{figure}